\documentclass[10pt]{article} 
\usepackage[preprint]{tmlr}

\usepackage{amsmath,amsfonts,bm}

\def\eqref#1{equation~\ref{#1}}

\def\1{\bm{1}}

\DeclareMathAlphabet{\mathsfit}{\encodingdefault}{\sfdefault}{m}{sl}
\SetMathAlphabet{\mathsfit}{bold}{\encodingdefault}{\sfdefault}{bx}{n}

\usepackage{hyperref}
\usepackage{url}

\usepackage[utf8]{inputenc} 
\usepackage[T1]{fontenc}    
\usepackage{hyperref}       
\usepackage{url}            
\usepackage{booktabs}       
\usepackage{amsfonts}       
\usepackage{nicefrac}       
\usepackage{microtype}      
\usepackage{xcolor}         
\usepackage{subcaption}
\usepackage{wrapfig}

\usepackage{amsmath}
\usepackage{algorithmic}
\PassOptionsToPackage{warn}{textcomp}  
\usepackage{textcomp}                  
\usepackage{times}                     
\usepackage{tabu}                      

\DeclareMathAlphabet{\mathcal}{OMS}{cmsy}{m}{n}
\usepackage[font={scriptsize,sf}]{caption}
\usepackage{multirow}
\usepackage{paralist}
\usepackage{subcaption}
\usepackage{cleveref}
\usepackage{comment}
\usepackage{sidecap}
\usepackage{stackengine}
\usepackage{graphicx}
\usepackage{enumitem}
\usepackage{float}

\title{GNNAnatomy: Rethinking Model-Level Explanations \\for Graph Neural Networks}

\author{\name Hsiao-Ying Lu \email hyllu@ucdavis.edu \\
      \addr University of California, Davis
      \AND
      \name Yiran Li \email ranli@ucdavis.edu \\
      \addr University of California, Davis
      \AND
      \name Ujwal Pratap Krishna Kaluvakolanu Thyagarajan \email ujwkal@ucdavis.edu\\
      \addr University of California, Davis
      \AND
      \name Kwan-Liu Ma \email klma@ucdavis.edu\\
      \addr University of California, Davis}

\begin{document}

\maketitle

\begin{abstract}
    Graph Neural Networks (GNNs) achieve state-of-the-art performance across graph-based tasks but remain difficult to interpret.
In this paper, we revisit foundational assumptions underlying model-level explanation methods for GNNs—namely: (1) maximizing classification confidence yields representative explanations, (2) a single explanation suffices for an entire class of graphs, and (3) explanations are inherently trustworthy.
We identify key pitfalls resulting from these assumptions: methods that optimize for classification confidence may overlook partially learned patterns; topological diversity across graph subsets within the same class is often underrepresented; and explanations alone offer limited support for building user trust when applied to new datasets or models.
To address these issues, we introduce GNNAnatomy, a distillation-based framework designed to generate explanations while avoiding these pitfalls.
GNNAnatomy first characterizes graph topology using graphlets—a set of fundamental substructures. We then train a transparent multilayer perceptron (MLP) surrogate to directly approximate GNN predictions based on the graphlet representations. By analyzing the weights assigned to each graphlet, we identify the most discriminative topologies, which serve as proxies for GNN explanations.
To account for structural diversity within a class, GNNAnatomy generates explanations at the required granularity through an interface that supports human-AI collaboration. This interface helps users identify subsets of graphs where distinct critical substructures drive class differentiation, enabling multi-grained explanations.
Additionally, by enabling interactive exploration and linking explanations back to input graphs, the interface fosters greater transparency and trust.
We evaluate GNNAnatomy on both synthetic and real-world datasets and demonstrate its advantages through quantitative metrics and qualitative comparisons with state-of-the-art model-level explainable GNN methods.

\end{abstract}

\newcommand{\CaptionTimeCost}{The time cost of each process in GNNAnatomy. }

\newcommand{\CaptionAlignments}{Alignment evaluation results reporting (1) the behavioral alignment between $MLP_{student}$ and $GCN_{teacher}$, and (2) the variance explained by the first principal component (PC1) in our projection map. Higher values indicate stronger alignment and better retention of data variability, respectively. }

\newcommand{\CaptionMMD}{The Maximum Mean Discrepancy (MMD) between the distributions of real graph embeddings and explanation embeddings, both obtained from the trained $GCN_{teacher}$. The consistently low observed MMD values indicate that the explanation embeddings are well-aligned with the distribution of real graph embeddings. Notably, MMD scores highlighted in red fall below the permutation-based null MMD range, suggesting a level of similarity that is statistically greater than expected under the null hypothesis.}

\newcommand{\CaptionProjectionMap}{Two selection of graphs, positioned on the left and right, indicate that different substructures within the magenta-colored class are necessary for the GNN to effectively differentiate this class.}

\newcommand{\CaptionInteractiveEval}{The representative visualizations display graphs from different classes, with the generated explanation highlighted in opaque black. This enables humans to connect the substructure explanation to the overall graph topology and interactively assess the quality of the explanation.}

\newcommand{\CaptionQual}{Model-level explanations generated by GNNAnatomy and other state-of-the-art methods. For each class in a dataset, the left is an example graph and the right is the explanation. We also provide the motifs attached to Barabási–Albert (BA) graphs in BA-2Motif dataset.}

\newcommand{\CaptionQuan}{The fidelity evaluation results for each model-level explanation corresponding to each class. The bolded text highlights the our contrasting explanation for each dataset.}

\newcommand{\CaptionKDworkflow}{The workflow of our knowledge distillation process and the formation of model-level explanations are illustrated as follows. The left section represents the training phase, where the bottom branch trains the $GCN_{teacher}$ model, while the top branch contains modules essential for training $MLP_{student}$. The right section depicts the inference phase, enabling the identification of the model-level explanation for class $c_i$ as the graphlet ($g$) that contributes the highest activation. The red text highlights the graphlet frequency vectors (GF) associated with graphs (G) of class $c_i$, while the red boxes indicate the weight and bias learned by $MLP_{student}$ for class $c_i$. }

\newcommand{\CaptionGraphletFull}{Graphlets with three nodes $(g_1, g_2)$, four nodes $(g_3-g_8)$, and five nodes $(g_9-g_{29})$.}

\newcommand{\CaptionMUTAGgraphs}{The top 10 graphs represent the Mutagen class in the MUTAG dataset, while the bottom 10 correspond to the Nonmutagen class. Both classes exhibit the common mutagenic substructure traits—a carbon ring attached to a functional group such as $NO_2$ or $NH_2$. The key topological distinction between them lies in the occurrence frequency of these mutagenic traits, which are noticeably more prevalent in the Mutagen class.}

\newcommand{\CaptionRBmulGrained}{This figure presents Discussion graphs from the Reddit-Binary dataset, illustrating the topological inconsistency within the class. The top 10 graphs are drawn from the left subset in \autoref{fig:projectionMap}, characterized by highly branching substructures. In contrast, the bottom 10 graphs are from the right cluster, where centralized substructures with significantly fewer branches dominate. This variation suggests that, for graphs in the latter subset, the GNN must rely on alternative structural features beyond the branching pattern to differentiate between classes.}

\newcommand{\CaptionRBfid}{This figure illustrates the structural changes in graphs after removing edges that participate in the corresponding model-level explanation. The top row shows the structural changes of a graph from the Discussion class in the Reddit-Binary dataset across different modification ratios. Similarly, the bottom row presents the structural variations of a graph from the Question-Answer class under the same modification conditions.}

\newcommand{\CaptionSubgraphMislead}{An example of how a subgraph explanation can be misleading due to imprecise.}

\newcommand{\CaptionGranularityInsufficient}{An example showing mutagens may exhibit different mutagenic compounds that distinguish them from non-mutagens.}

\newcommand{\CaptionGraphlet}{Six graphlet types with four nodes.\vspace{-0.1in}}

\newcommand{\CaptionGNNLayer}{The process of relational information aggregation in GNNs.}

\newcommand{\CaptionUI}{
The interface of GNNAnatomy, a screen capture of generating and reviewing explanations for the GNN behavior on a selected group of graphs from the Reddit-Binary dataset. In Column a, (a1) displays the legends for color encoding; (a2) shows the projection map visualizing the proximity of graphs, guiding users to select a group of graphs for which explanations are generated; and (a3) allows users to refine their selection based on class. After clicking the ``generate explanation'' button, Column b displays graphlets ranked according to their relevance to the GNN’s classifications; (b1) depicts each graphlet, while (b2) presents a class-specific histogram showing the distribution of graphs based on graphlet frequency. Subsequently, users can evaluate the relevance of a specific graphlet by selecting it, as shown in column (c), where $graphlet_9$ is chosen; (c1) displays the correlation pattern between the activation produced by $graphlet_9$’s frequency and the GNN's classification probability, while (c2) illustrates the changes in classification confidence for each graph after removing edges in $graphlet_9$ substructures. Users may lasso-select graphs in (c1) to see the overall topological impact of the chosen substructure, $graphlet_9$. In Column d, (d1) and (d2) then display representative graphs from two different classes, providing the connection between $graphlet_9$ substructure and the overall topology. Users can examine the occurrences of $graphlet_9$ within each graph by clicking the ``graphlet highlight'' button.
}

\newcommand{\CaptionRBCompare}{Reddit-Binary dataset. The class motifs inferred by GNNExplainer and substructure explanations generated GNNAnatomy.}

\newcommand{\CaptionRBoverlap}{
(a) illustrates the selection of a group of graphs with similar topologies, aimed at observing subtle topological differences between the two classes in the Reddit-Binary dataset. (b) depicts the top-ranked $graphlet_{19}$ and the graph distribution over the its frequency. (c) shows five examples where $graphlet_{19}$ locates in a representative Question-Answer graph.
}

\newcommand{\CaptionMutag}{Starting from (a), a group of graphs is selected to investigate the key structural differences between mutagens and non-mutagens. In (b), each graphlet is ranked according to its relevance to the GNN's classifications. After selecting the top-ranked ring graphlet, (c) shows a scatterplot illustrating the correlation between the frequency of the ring graphlet and the classification probability. With the graphs having contrasting graphlet frequencies chosen, (d1) and (d2) reveal the overall topology reflecting this frequency difference and highlight the ring graphlet within their topology. 
}

\newcommand{\CaptionBA}{
In (a1), a group of graphs is selected to examine the subtle structural differences between House class graphs and Cycle class graphs. (a2) highlights the strong relevance of the triangle substructure. Upon selecting the triangle graphlet, (a3) displays a scatterplot showing the strong correlation between the frequency of the triangle graphlet and the classification probability. Additionally, (a4) shows a decrease in the classification probability of House graphs after the triangle graphlet is removed. (b) illustrates how house motifs are attached to BA base graphs, with the house motifs highlighted.
}

\newcommand{\CaptionBACompare}{
BA-2Motif dataset. The ground truth motifs and the explanations generated by a baseline and GNNAnatomy.
}

\newcommand{\CaptionMutagCompare}{MUTAG dataset. The ground truth mutagenic compounds, which include a fused ring, a 6-node ring structure, and $NO_2$ (from right to left), along with the explanations provided by the baseline methods and GNNAnatomy.
}
\newcommand{\ProbDensFunc}{f}
\newcommand{\ProbDensFuncAll}{\ProbDensFunc_\mathrm{all}}
\newcommand{\ProbDensFuncClassZero}{\ProbDensFunc_\mathrm{0}}
\newcommand{\ProbDensFuncClassOne}{\ProbDensFunc_\mathrm{1}}

\newcommand{\nInsts}{n}
\newcommand{\nInstsClassZero}{\nInsts_\mathrm{0}}
\newcommand{\nInstsClassOne}{\nInsts_\mathrm{1}}

\newcommand{\DensRatioFunc}{g}
\newcommand{\DensRatioFuncClassZero}{\DensRatioFunc_\mathrm{0}}
\newcommand{\DensRatioFuncClassOne}{\DensRatioFunc_\mathrm{1}}

\newcommand{\Point}{\mathbf{p}}

\newcommand{\TransRate}{r}
\section{Introduction}
\label{sec:intro}
Graph Neural Networks (GNNs) have become increasingly prominent for their superior performance in graph data tasks, impacting various domains such as social network analysis (\cite{dinh2021social, abbas2021social, min2021stgsn, tan2019deep}) and biological science research (\cite{li2021graph, muzio2021biological, jin2021application}). GNNs convert complex graph structures into vector representations by aggregating relational information from neighboring nodes (\cite{ying2018hierarchical, kipf2016semi}), capturing intricate topological features that enable accurate predictions or classifications.
While achieving high accuracy is crucial, interpretability is equally important (\cite{pope2019explainability, yuan2022explainability}). Understanding the topological traits that GNNs rely on for predictions empowers decision-makers to optimize outcomes in complex, real-world scenarios. Moreover, uncovering potential spurious associations captured by GNNs allows machine learning researchers to diagnose and refine model behavior. Therefore, integrating interpretability into GNNs is essential for fostering trust and transparency.

Existing methods for explaining GNNs can be categorized into instance-level and model-level approaches. Instance-level explanations aim to clarify the GNN’s behavior on a single graph instance (\cite{ying2019gnnexplainer, zhang2021relex, vu2020pgm}). While effective for understanding specific predictions, these explanations often lack generalizability across different graphs, limiting their ability to provide a holistic understanding of the GNN behavior. Consequently, recent work has shifted towards model-level explanations (\cite{yuan2020xgnn, wang2022gnninterpreter, chen2024d4explainer}), which seek to uncover how GNNs differentiate a class of graphs. However, despite their growing prominence, existing model-level methods rest on under-examined assumptions. 

\textbf{Model-level explanations.} First, model-level explanations are typically generated by optimizing the GNN’s classification probability to synthesize an explanatory graph pattern for each class. However, these explanations risk falling out of the distribution of the original graphs. Even with constraints to enforce distributional similarity, they may still fail to represent graphs that the GNN only partially learned, as such graphs yield lower prediction confidence. As a result, explanations optimized for maximum probability primarily apply only for the graphs that GNN has fully learned.
Second, existing model-level approaches often aim to explain the GNN’s behavior on an entire class using a single explanation. However, even within the same class, graphs can exhibit incoherent topological traits that vary across different subsets. Relying on a single explanation for an entire class overlooks this variability, reducing granularity and specificity.
Third, most model-level explanations rely on the raw GNN prediction probability as a metric. This offers limited insight into the sufficiency and necessity (\cite{pope2019explainability, hooker2019benchmark}) of the generated explanations. Additionally, many approaches depend on another black-box model to generate explanations (\cite{chen2024d4explainer, yuan2020xgnn}). As highlighted by Wang et al. (\cite{wang2022gnninterpreter}), this approach merely shifts the trust issue from the GNN to its explainer, rather than resolving it.

\textbf{Presented approach.} To address these limitations and avoid the assumption-driven pitfalls, we introduce a distillation-based approach, \textbf{\textit{GNNAnatomy}}, which explains the GNN (teacher model) by interpreting a student model trained to directly approximate its behavior. 
The distillation begins by representing each graph as a vector of graphlet occurrence frequencies, offering an anatomical breakdown of its fundamental substructures. A single-layer MLP then takes these graphlet frequency vectors as input and is trained to align its outputs with the GNN’s probability predictions. The model-level explanation for each class is identified as the graphlet whose frequency contributes the most to the MLP’s activation. Finally, the explanation’s sufficiency and necessity are assessed using fidelity metrics. 
To account for structural diversity within a class, GNNAnatomy integrates a visual analytics interface to enhance human-AI collaboration by assisting users in identifying subsets of graphs that exhibit incoherent structural traits through visualizing the relationship between GNN classifications and graphlet frequencies. This enables the discovery of contrasting substructures necessary for class differentiation within each subset, which extends beyond traditional model-level explanations to enable explanations at a user-specified, required  granularity. Additionally, the interactive interface empowers users to explore the relationships between the original graph structure and their corresponding explanations, fostering greater trust in the generated insights.


We demonstrate the effectiveness of GNNAnatomy on real-world and synthetic graph datasets across different domains, comparing our explanations to those generated from state-of-the-art model-level explainable GNN methods. These case studies showcase GNNAnatomy's ability to capture compact and intrinsic topological differences between classes. Overall, \textbf{our contributions include}: 

\vspace{0.05in}
\begin{itemize}[nosep, leftmargin=*]
    \item Through distillation, we ensure (1) the alignment between the distribution of generated explanations and that of the real graphs, and (2) the explanations' representativeness of the GNN’s behavior across all graphs, including those with varying levels of classification confidence.
    \item Through human-AI collaboration, we enable the identification of explanations at flexible levels of required granularity. 
    \item Through fidelity metrics, we evaluate model-level explanations in terms of their sufficiency and necessity.
    \item Through employing an interpretable explaining model and allowing free exploration of the connections between explanations and real graphs via our interface, we enhance the trustworthiness of our explanations.
\end{itemize}

\section{Background}
\label{sec:background}


    \begin{wrapfigure}{r}{0.5\textwidth}
        \begin{center}
        \includegraphics[width=0.48\textwidth]{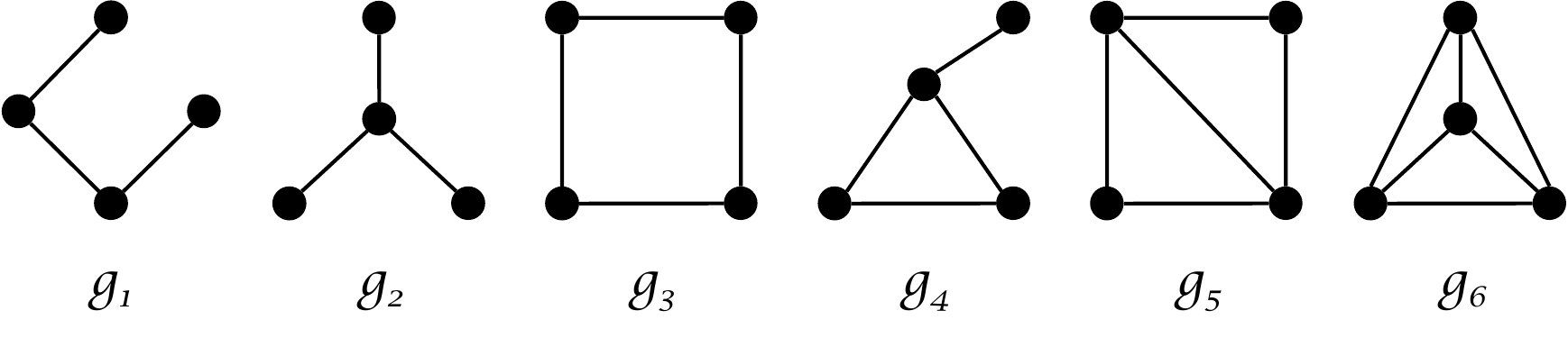}
        \end{center}
        \caption{\CaptionGraphlet{}}
        \label{fig:graphlets}
    \end{wrapfigure}
    \textbf{Graphlets and Graphlet Frequencies. }
    \label{sec:background_graphlet}
    Graphlets are small connected non-isomorphic graphs (\cite{prvzulj2007biological, faust2010triads}). Given a specific number of nodes in a graphlet, there is a fixed number of graphlet types. 
    For example, we show all 6 graphlet types with 4 nodes in \autoref{fig:graphlets}. Graphlets have been used in both analysis and visualization of graphs/networks (\cite{Ugander,Kwon2018}).
    
     

    The relative frequencies of different graphlet types can be used to characterize the topology of a graph. If we would use the 6 graphlet types $g_1, g_2, \cdots, g_{6}$ in \autoref{fig:graphlets} to characterize a set of graphs, a 6D vector of graphlet frequencies $\mathbf{f}_G$ is extracted to represent the graphlet distribution in each graph $G$:
    \begin{equation}
        \mathbf{f}_G = (f_{g_1}, f_{g_2}, \cdots, f_{g_{6}}),
    \end{equation}
    where 
    \begin{equation}
        f_{g_k} = \frac{\#(g_k \subseteq G)}{\sum_{i=1}^{6} \#(g_i \subseteq G)}.
    \end{equation}

    \textbf{Graph Neural Networks. }
    \label{sec:backgroundGNN}
    Graph Neural Networks (GNNs) learn structure-aware node representations by iteratively gathering information from a node's neighbors. In each iteration or a layer $l$ in a GNN, a node $v$ aggregates information from its directly connected neighbors by a function. This aggregated information is then integrated with the node's current representation $h_v^{(l-1)}$ by another function, which can be presented as $h_v^{(l)} = Update(h_v^{(l-1)}, Aggregate_{(l)}(h_{u}^{(l-1)}, \forall u\in \mathcal{N}(v)))$, where $\mathcal{N}(v)$ is the neighbors of node $v$ and $h_v^{(0)}$ is the node features of $v$. 
    Any GNNs following this message-passing mechanism can be explained by GNNAnatomy.
\section{Related Works}
\label{sec:related}

    \textbf{Explainable GNN Methods. }
        As outlined in a recent survey (\cite{yuan2022explainability}), explainable GNN methods generally produce either instance-level or model-level explanations. Instance-level explanation identifies a subgraph most relevant to the GNN's prediction for a graph instance (\cite{ying2019gnnexplainer, yuan2021explainability, funke2020hard, schlichtkrull2021interpreting, wang2020causal, huang2022graphlime, zhang2021relex, vu2020pgm}). Yet, such explanations may not be applicable to other graph instances, limiting their ability to provide a general understanding of GNN behavior.
        
        Consequently, recent research has shifted toward model-level explanations, which aim to highlight distinguishing substructures for each class. Existing methods, such as XGNN and GNNInterpreter (\cite{yuan2020xgnn, wang2022gnninterpreter}), achieve this by training graph generators to synthesize graphs that maximize the GNN’s classification confidence for each class. D4Explainer (\cite{chen2024d4explainer}) further enhances the in-distribution generation through reverse sampling and a denoising model. 
        However, these generative approaches face two key limitations. First, ensuring distributional alignment between explanations and real graphs is difficult when synthetic generation is involved. While additional constraints can improve distributional similarity, a deeper issue remains: the optimization objective centers on maximizing prediction confidence, not explicitly approximating the full output behavior of the GNN. As a result, a critical subset of the data is underrepresented — graphs where the GNN makes lower-confidence, partially learned predictions are unlikely to be captured in explanations driven solely by high-confidence optimization.
        Addressing this gap is crucial for explanations to faithfully represent the full spectrum of GNN behavior across diverse graph instances.
        
        Finally, even model-level approaches could fall short in capturing the topological diversity present within a class. Different subsets of graphs can rely on distinct structural features to drive class differentiation, yet existing methods typically overlook this variability. To effectively reflect GNN behavior, it is essential to offer explanations at flexible levels of required granularity.

    \textbf{Evaluation of GNN Explanations. }
    \label{sec:relatedEval}
        To evaluate the performance of instance-level explanations, accuracy-based measures, such as F1 or ROC-AUC scores (\cite{ying2019gnnexplainer, sanchez2020evaluating}) rely on the availability of ground truth, Fidelity metrics (\cite{pope2019explainability, hooker2019benchmark}) measure the changes in GNN prediction when perturbations are made, Sparsity metrics (\cite{pope2019explainability}) assess how compact the explanations are.
        However, most model-level explanations are evaluated solely based on the GNN’s prediction probability (\cite{yuan2020xgnn, wang2022gnninterpreter}), offering limited validation of the accuracy and fidelity of the generated explanations.
        Additionally, many explainable GNN methods rely on another black-box model to generate explanations, which merely shifts the trust issues from the GNN to its explainer, as highlighted by GNNInterpreter (\cite{wang2022gnninterpreter}). A more effective approach would be to offer transparent and ample evidence that enable users to reason about the GNN's behavior and arrive at a trustworthy explanation.

    \textbf{Visual Analytics for Explaining GNNs. }
        To enhance the trustworthiness of generated explanations, visual analytics (VA) has proven to be an effective technique, allowing users to interactively explore and interpret complex data. 
        Recent surveys (\cite{la2023state, wang2024visual}) highlight a growing interest in visual analytics (VA) within explainable ML research, though only a small portion specifically targets GNN explainability. 
        However, most methods (\cite{li2018embeddingvis, heimerl2020embcomp, liu2022visualizing, jin2022gnnlens}) fall short of offering intuitive explanations that allow users to quickly grasp the topological or semantic features key to a GNN’s predictions. Instead, these VA tools provide ample evidence but rely on users to extract meaningful insights and construct their own explanations. This can lead to inconsistent interpretations, making it difficult to assess the validity of the explanations.

\section{GNNAnatomy}
\label{sec:method}
    As outlined in \autoref{sec:intro} and \autoref{sec:related}, explanations generated by existing model-level methods suffer from three assumption-driven pitfalls: (1) limited representativeness of partially learned graphs, (2) insufficient granularity due to structural variability within each graph class, and (3) constrained trustworthiness stemming from black-box explainers and the insufficient evaluation relying solely on prediction confidence.
    In this section, we describe our solutions for addressing these challenges. 
    Since explaining GNNs is intricately tied to understanding graph structures rather than just node or edge features, as also noted by \cite{luo2020parameterized}. Additionally, uncovering relevant features for non-graph neural networks is a well-studied field (\cite{fong2017interpretable, ribeiro2016should, lundberg2017unified}) and GraphLime (\cite{huang2022graphlime}) also explored to explain GNN's behavior using only neighboring nodes' features.
    Thus, our work is deliberately scoped to focus on structural explanations that are most relevant to GNN behavior in graph classification tasks. Feature attribution is deferred to complementary methods, as discussed above, allowing for modular integration while avoiding the conflation of structural and feature-based interpretability.

          
    \subsection{Knowledge Distillation: Direct Model Behavior Alignment}   
    \label{sec:method_KD}
        Surrogate-based approaches distill GNN knowledge into simpler, interpretable models, using explanations from these models as proxies for GNN behavior. However, such applications have been largely limited to instance-level explanations (\cite{huang2022graphlime, zhang2021relex, vu2020pgm, pereira2023distill}). In contrast, model-level methods have seldom leveraged the interpretability and alignment benefits of distillation. In this work, GNNAnatomy advances this paradigm by combining graphlet-based representations with a transparent surrogate model to directly approximate GNN behavior while preserving interpretability.
        
    \textbf{Graphlet Frequency: Topological Summary. }
    \label{sec:method_graphlet}
        Since GNNs aggregate node information through direct connections, they inherently integrate messages contained in small connected substructures, with their size and diameter determined by the number of layers in the GNN architecture. To make these captured substructures explicit and enhance interpretability, we first analyze the substructures present in a graph. 
        
        Our method employs graphlets to characterize each graph's topology into graphlet frequency vector, as outlined in \autoref{sec:background}. For any graph $G$, we concatenate the 3-node (2D), 4-node (6D), and 5-node (21D) graphlet frequency vectors into a 29D graphlet representation. The full list of the default 29 graphlets can be found in Appendix C.
        The decision to use graphlets of up to 5 nodes is motivated by both interpretability and practical considerations. From an interpretability standpoint, incorporating larger graphlets—such as those with 6 nodes—would introduce 112 additional patterns, making meaningful human interpretation and differentiation infeasible. GNNAnatomy remains flexible, however, and can accommodate additional substructures if specific domain needs arise.
        From a practical perspective, 3- to 5-node graphlets already capture a diverse set of small structural patterns that can compose more complex motifs. Moreover, most GNN architectures use 4 or fewer layers, meaning that structural information within a 4-hop neighborhood is repeatedly aggregated and thus most strongly represented in the learned embeddings. Importantly, the largest diameter among 5-node graphlets is 4 (e.g., the 5-node chain), ensuring that the most prominently expressed substructures in typical GNNs are already captured by our selected graphlets.
        
        In the actual computation of graphlet frequencies, it is too time-consuming to count the exact occurrence of each graphlet because comparing all possible connected subgraphs is an NP-hard problem. To address this, we apply Louvain community detection (\cite{blondel2008fast}) to partition large graphs into smaller, more manageable communities. We then compute the occurrences of each graphlet type within these detected communities of $G$. The total frequency of a graphlet in $G$ is calculated based on its aggregated occurrences and the total number of connected subgraphs considered across all communities.

          
    \begin{figure*}
        \centering
        \includegraphics[width=\linewidth]{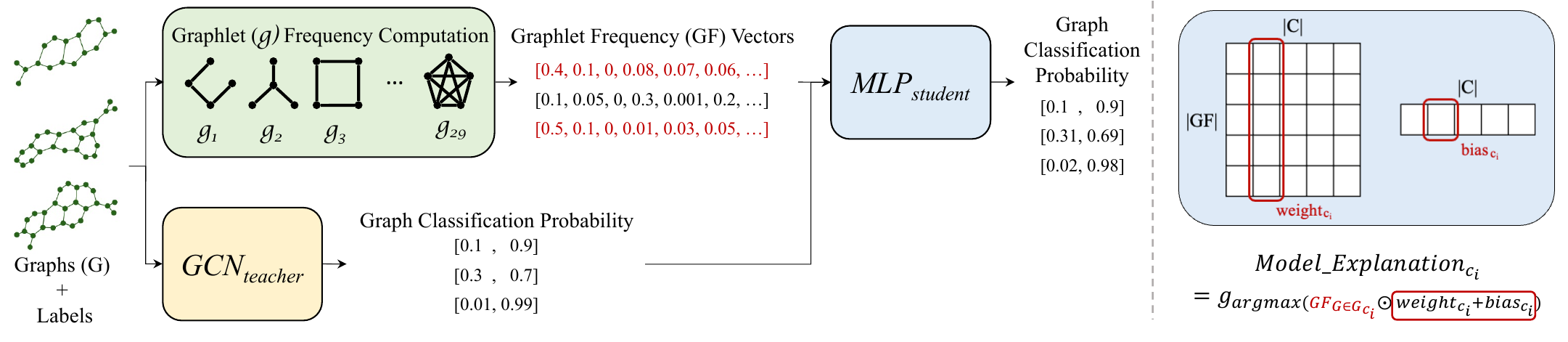}
        \caption{\CaptionKDworkflow{}}
        \label{fig:KDworkflow}
    \end{figure*}
    
    \textbf{Multilayer Perceptron: GNN Prediction Approximation from Graphlet Frequencies. }
    \label{sec:method_MLP}
        The computed graphlet frequency vectors are fed into a multilayer perceptron ($MLP_{student}$), which is trained to directly approximate the probability output of the trained GNN. The architecture of $MLP_{student}$ consists of a single fully connected layer followed by a softmax function, with no nonlinear activation functions involved. This design promotes model transparency by relying solely on a linear combination of graphlet frequencies to approximate the GNN’s behavior, achieved by aligning their output spaces.
        
    \textbf{MLP Activation Analysis: Model-Level Explanations.}
    \label{sec:method_EXP}
        Leveraging the interpretability of our MLP design, we now describe how explanations are derived from the learned parameters in $MLP_{student}$. The weight matrix has a shape of $|GF| \times |C|$, and the bias vector has a shape of $|C|$, where $|GF|$ represents the number of graphlets included in the frequency vector, and $|C|$ is the number of classes.
        To determine the model-level explanation for class $c_i$, we identify the graphlet that produces the highest activation weighted by the weight and bias of class $c_i$ among graphs belonging to class $c_i$ ($G_{c_i}$). This graphlet is then considered the model-level explanation for class $c_i$. 
        \begin{equation}
        \label{eqt:expIndex}
            \Phi(c_{G}, c_{wb}) = 
            \frac{1}{|G_{c_{G}}|}\underset{G \in G_{c_{G}}}{\sum}GF_{G} \odot weight_{c_{wb}} + bias_{c_{wb}}
        \end{equation}
        \begin{equation}
        \label{eqt:modelExp}
            Model\_Explanation_{c_i} = graphlet_{argmax(\Phi(c_i, c_i))}
        \end{equation}, where $G_{c_G}$ is the graphs from class $c_G$ and $GF_G$ is the graphlet frequency vector of graph $G$. The entire workflow of our distillation process is illustrated in \autoref{fig:KDworkflow}.

        After deriving the primary model-level explanation using \autoref{eqt:expIndex} and \autoref{eqt:modelExp}, we compute the Pearson correlation between its weighted frequency and those of all other graphlets. Graphlets with a correlation greater than 0.5 are identified as auxiliary explanations, as they frequently co-occur with the primary pattern. To reduce redundancy, any graphlet that is a strict subgraph of the primary graphlet (e.g., omitting a 4-node star when the 5-node star is identified as primary) is excluded. The combination of the primary graphlet and its auxiliaries constitutes the final model-level explanation.

    \textbf{What this approach solves.}
    \label{sec:method_KDSolve}
        The existing alignment issues stem from how prior model-level methods link their generative explanations to GNN behavior—by synthesizing graphs that maximize classification probability.
        GNNAnatomy addresses this in two ways. First, it extracts graphlet occurrences directly from the original graphs, ensuring explanations are grounded in patterns actually present in the dataset, thereby preserving distributional similarity. Second, it trains a transparent MLP to directly approximate GNN behavior and identifies the graphlets whose frequencies best align the GNN classification probabilities. This approach captures core structural signals that are representative across the full spectrum of graphs, not just those with high classification confidence.
        Additionally, the transparency of $MLP_{student}$ enhances the trustworthiness of the explanations, while the sparsity—measured by node coverage in the original graph—improves interpretability by grounding explanations in compact, comprehensible substructures. Since we examine all substructures from 3-node (the smallest possible) to 5-node graphlets, our explanations are inherently minimal in size.

    \subsection{Visual Analytics: Human-AI Teaming for Multi-Grained Explanations}
    \label{sec:method_VA}
        As discussed in \autoref{sec:related}, a model-level explanation identifies a distinguishing substructure for a class, but graphs within this class may exhibit inconsistent topological traits. This inconsistency could lead the GNN to rely on a different substructure to differentiate classes for graphs that show fewer of the structural traits associated with the model-level explanation. Therefore, it is crucial to capture this granularity by (1) identifying subsets of graphs that exhibit such inconsistencies and (2) pinpointing the substructures used to differentiate classes within these subsets. 
        Additionally, all explainable GNN methods include qualitative evaluations where the generated explanations are presented alongside real graph instances. Without the ability to review explanation-graph pairs for datasets beyond those used in published evaluations, it is difficult to assess their broader applicability and makes it challenging to develop trust in their reliability. 
   
        In response to these limitations, our visual analytics interface serves as a pivotal tool, guiding users through visualizations to identify subsets of graphs that rely on different substructures for class differentiation and enabling users to freely explore the connections between explanations and real graph instances, fostering greater trust in the results generated by GNNAnatomy. In this section, we describe the key visualization component, while a comprehensive introduction to our interface is provided in the Appendix A. 
        
    \begin{wrapfigure}{R}{0.63\textwidth}
        \begin{center}
        \vspace{-0.25in}
        \includegraphics[width=0.61\textwidth]{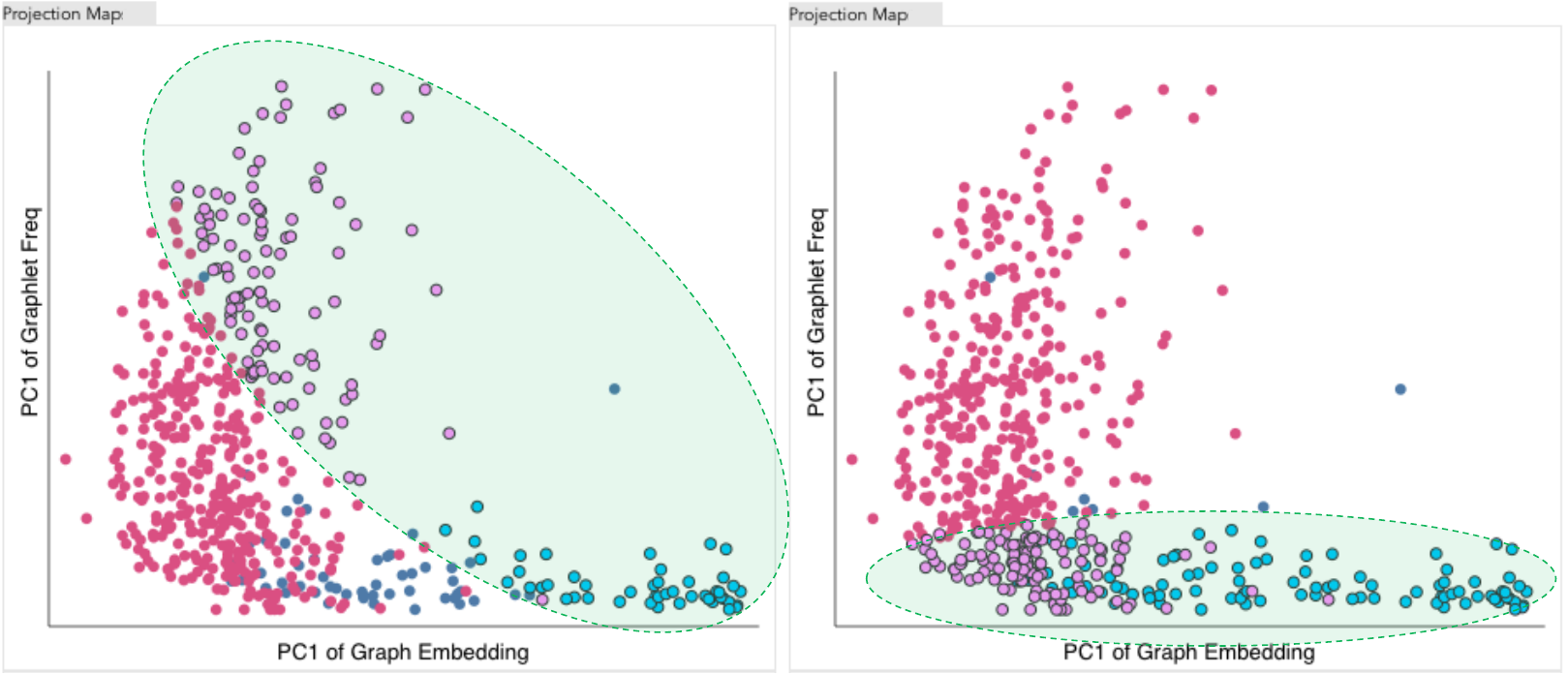}
        \end{center}
        \caption{\CaptionProjectionMap{}}
        \label{fig:projectionMap}
    \end{wrapfigure}
        
    \textbf{Projection Map: Graph Subset Selection.}
    \label{sec:method_PM}
        The projection map, as shown in \autoref{fig:projectionMap}, demonstrate how to reveal subsets of graphs that require different substructures for GNNs to differentiate between classes in a graph-level binary classification task. This visualization projects all graphs (represented as dots in this scatterplot) from the analyzed dataset. The x-axis represents the first principal component (PC1) of the graph embedding generated by the trained GNN. The proximity along the x-axis indicates how similarly the GNN perceives the graphs; graphs closer together are considered more similar in terms of the topological traits captured to generate their embeddings. The y-axis represents the PC1 of the graphlet frequency vectors, with proximity along this axis reflecting the similarity in graph topologies as characterized by the graphlets. Graphs closer on the y-axis have more similar topologies. 
        
        Notably, the pink class of graphs is scattered along the y-axis, indicating that the topological traits within this class are incoherent. Consequently, the diagonal (\autoref{fig:projectionMap}-left) and horizontal (\autoref{fig:projectionMap}-right) subsets of graphs likely require different substructures for effective class differentiation.
        Finally, the choice of using a linear dimensionality reduction method—principal component analysis (PCA) aligns with the linear design of $MLP_{student}$, ensuring that the separation and proximity captured by the linear projections can reflect the knowledge learned by $MLP_{student}$.

    \textbf{Multi-Grained Explanations.}
    \label{sec:method_mulG}
        Guided by the associations revealed in the projection map, user can lasso-select a subset of graphs, which enables the generation of multi-grained explanations in GNNAnatomy. This generation aims to provide the one substructure that contributes the most to the class differentiation within the selected subset. For example, after selecting a graph subset $k$ containing graphs from classes $c_0$ and $c_1$, we leverage the distilled global knowledge in $MLP_{student}$ and \autoref{eqt:expIndex} to identify the graphlet whose frequency contributes the most to increasing activation (weighted by the weight and bias of $c_0$) for class $c_0$ among graphs within subset $k$ and from class $c_0$, while simultaneously contributing the most to decreasing activation among graphs from class $c_1$. Conversely, when activating with the weight and bias of class $c_1$, we seek the graphlet whose frequency leads to the highest activation among graphs from $c_1$ and the lowest among graphs from $c_0$. We introduce the subset $k$ parameter into \autoref{eqt:expIndex}: 
        \begin{equation}
            \Phi_k(c_{G}, c_{wb}) = 
            \frac{1}{|G^k_{c_{G}}|}\underset{G \in G^k_{c_{G}}}{\sum}GF_{G} \odot weight_{c_{wb}} + bias_{c_{wb}}
        \end{equation}
        , where $G^k_{c_{G}}$ denotes the graphs from class $c_G$ and within subset $k$. Then, the contrasting explanation of subset $k$ is defined by
        \begin{equation}
        \label{eqt:expIndexGrained}
            m = argmax([\Phi_k(c_{0}, c_{0})-\Phi_k(c_{1}, c_{0})] + [-\Phi_k(c_{0}, c_{1})+\Phi_k(c_{1}, c_{1})])
        \end{equation}
        \begin{equation}
            Contrasting\_Explanation^k_{c_0, c_1} = graphlet_{m}
        \end{equation}
        The intuition behind this equation is to compare activations between graphs from classes $c_0$ and $c_1$ in order to identify the graphlet whose frequency most distinctly differentiates class-specific activations within the selected subset $k$. Throughout this paper, we refer to this graphlet as the contrasting explanation. Notably, compared to other methods that aim to generate multi-grained explanations (\cite{wang2021towards, yin2023train}), our contrasting explanation offers greater interpretability and provides finer-grained information. Further discussion can be found in the Appendix B.

\section{Experiments}
\label{sec:studies}
    We demonstrate the capabilities of GNNAnatomy through studies using both real-world and synthetic graph datasets. Additionally, the generated explanatory substructures are evaluated and compared to state-of-the-art model-level explainable GNN methods: XGNN (\cite{yuan2020xgnn}) and GNNInterpreter (\cite{wang2022gnninterpreter}) based on the best results reported in their papers.

    \subsection{Experimental Setup}
    
    \textbf{Real-world datasets.}
        \textit{Reddit-Binary} (\cite{yanardag2015deep}) contains 2000 graphs, each representing a Reddit thread labeled as either Question-Answer or Discussion based on user interaction patterns.
        \textit{MUTAG} (\cite{debnath1991structure}) contains 188 molecular graphs labeled by their mutagenic effect on Salmonella typhimurium.

    \textbf{Synthetic datasets.}
        \textit{BA-2Motif} (\cite{luo2020parameterized}) contains 1000 graphs generated from Barabási-Albert (BA) model, with half attached to a house motif and half to a 5-node cycle motif. 
        \textit{Shape} (\cite{wang2022gnninterpreter}) comprises graphs labeled as Lollipop, Wheel, Grid, Star, or Others, where the Others class contains randomly generated graphs without specific structure. Random noise is added to each graph to enhance model robustness.


    \textbf{Graph Classification Model.}
        We use the Graph Convolutional Network (GCN) (\cite{kipf2016semi}) as the GNN model to be explained on the selected datasets. 
        GCN is chosen because it shares the same learned weights for aggregating information from same-hop neighbors across the entire graph. This ensures that identical substructures are valued consistently, no matter where they appear within a graph.
        As discussed in \autoref{sec:method}, our work is deliberately scoped to focus on delivering structural explanations. Hence, we omit any node or edge attributes during training and instead initialize each node feature as a constant vector. 

        Our GCN architecture, $GCN_{teacher}$, consists of four graph convolutional layers followed by a fully connected layer for classification. The output from each convolutional layer, with a dimensionality of 20, is concatenated to create an 80-dimensional node embedding for each graph. All node embeddings within a graph are max-pooled to form an 80-dimensional graph embedding. 
        The number of GNN convolutional layers and the size of the included graphlets are interdependent, as discussed in \autoref{sec:method_KD}. Given that the largest diameter among the selected graphlets is 4 (e.g., the 5-node chain), a 4-layer GNN suffices to capture their topological information. To ensure a balanced representation across substructures of varying sizes, we concatenate the hidden outputs from all convolutional layers. This aggregation enables the final graph embeddings to reflect features from both low- and high-hop neighborhoods, mitigating potential overemphasis on larger graphlets in the deepest layer.

    \subsection{Results}
    \label{sec:caseStudy}
    \begin{figure*}[!ht]
        \centering
        \includegraphics[width=\linewidth]{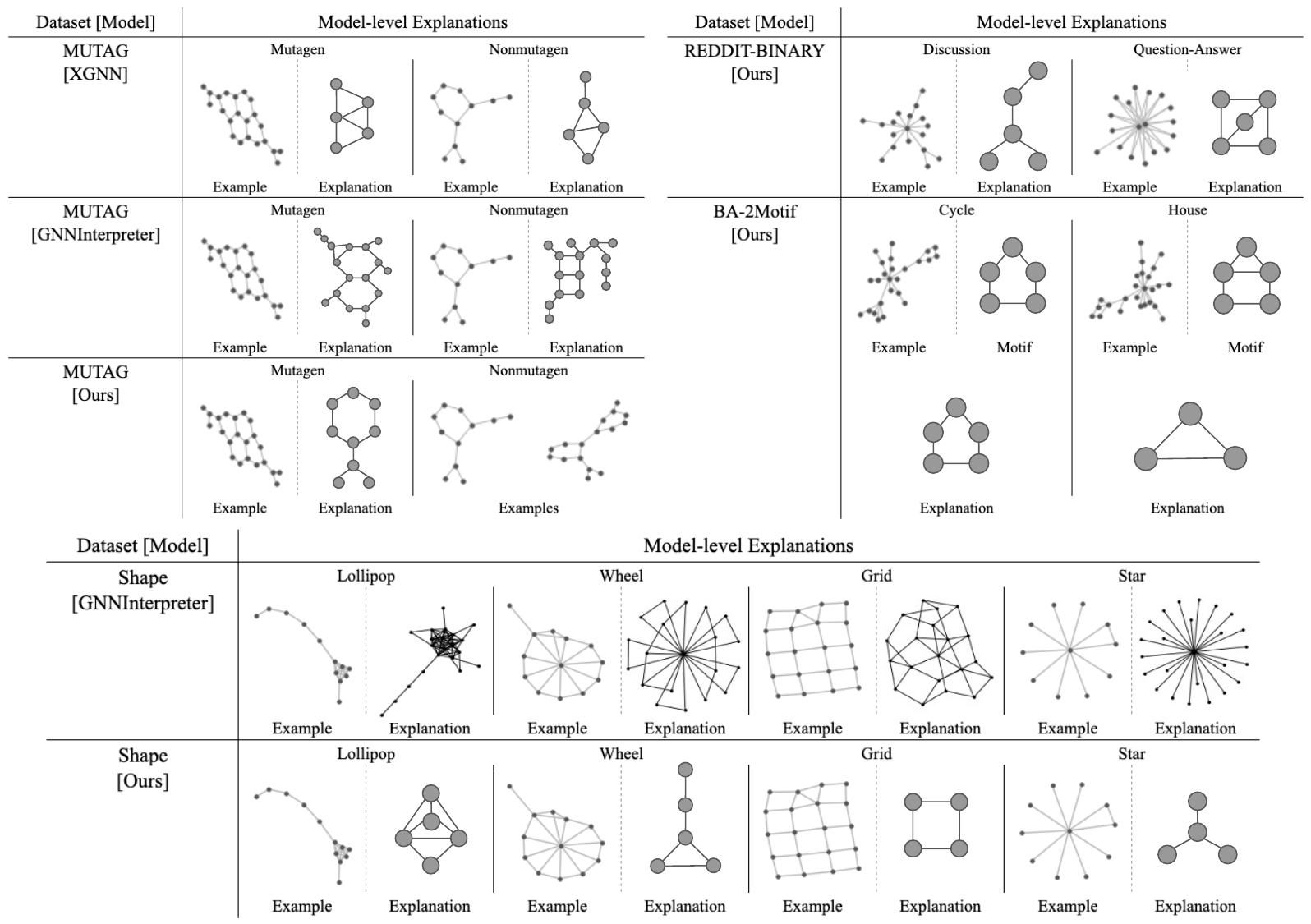}
        \caption{\CaptionQual}
        \label{fig:qual}
    \end{figure*}

        As shown in \autoref{fig:qual}, our explanations capture the fundamental structures of each class. We further discuss each dataset separately. 
        
        \textbf{Shape: Non-Binary Classification Explanations.} 
        \label{sec:ShapeStudy}
            Graphs in the Shape dataset are generated based on predefined rules. As illustrated in \autoref{fig:qual}, a Lollipop graph consists of a clique connected to a chain of nodes. Our explanation highlights the clique substructure with the top four nodes, representing the most elementary and essential component. Since any clique larger than four nodes contains a 4-node clique, this explanation captures the core distinguishing feature. The bottom node in the explanation may indicate that the clique extends beyond four nodes and connects to a chain structure. Similarly, our explanations for Grid and Star classes identify the fundamental structures that best represent these classes. For the Wheel class, the explanation reveals that GNN appears to be influenced by the introduced noise, and thus, incorporates a 2-node chain above the triangle. However, the triangle itself remains the most characteristic substructure within the Wheel class. 
            In contrast, GNNInterpreter produces explanations of arbitrary size. Although some may resemble individual input graphs, this randomness limits their ability to reflect the fundamental structural patterns shared across most or all graphs within a class.
            Additionally, its explanations for the Wheel and Grid classes appear inconsistent with the actual shapes of the original graphs, reducing their interpretability and alignment with the underlying graph structures.

        \textbf{BA-2Motif: Identification of Sparse Explanations.} 
        \label{sec:BAStudy}
            A BA-2Motif graph from one class differs from the other only by the specific motif attached to its randomly generated base graph. As illustrated in \autoref{fig:qual}, the two motifs—Cycle and House—differ by just one edge. Our explanations effectively capture that GNN perceives that a triangle alone can distinguish House graphs, whereas a complete 5-node cycle is required to confirm a graph belonging to the Cycle class. 
            Furthermore, because House motifs always connect to the base graph through a shared node between the triangle and rectangle, a GNN requires one fewer layer to aggregate information from a triangle substructure. This insight aligns with our explanations, highlighting both the efficiency of GNNs and a potential risk—an unseen graph with any triangle in its structure could be misclassified as a House graph.

        \textbf{MUTAG: Explanations via Substructure Concurrence.}
        \label{sec:MutagStudy}
            We incorporate a 6-node ring graphlet into this study for known mutagenic structure (i.e., 6-atom carbon ring).
            Additionally, we apply \autoref{eqt:expIndexGrained} to the entire set 
            \begin{wrapfigure}{r}{0.5\textwidth}
                \begin{center}
                \vspace{-0.1in}
                \includegraphics[width=0.49\textwidth]{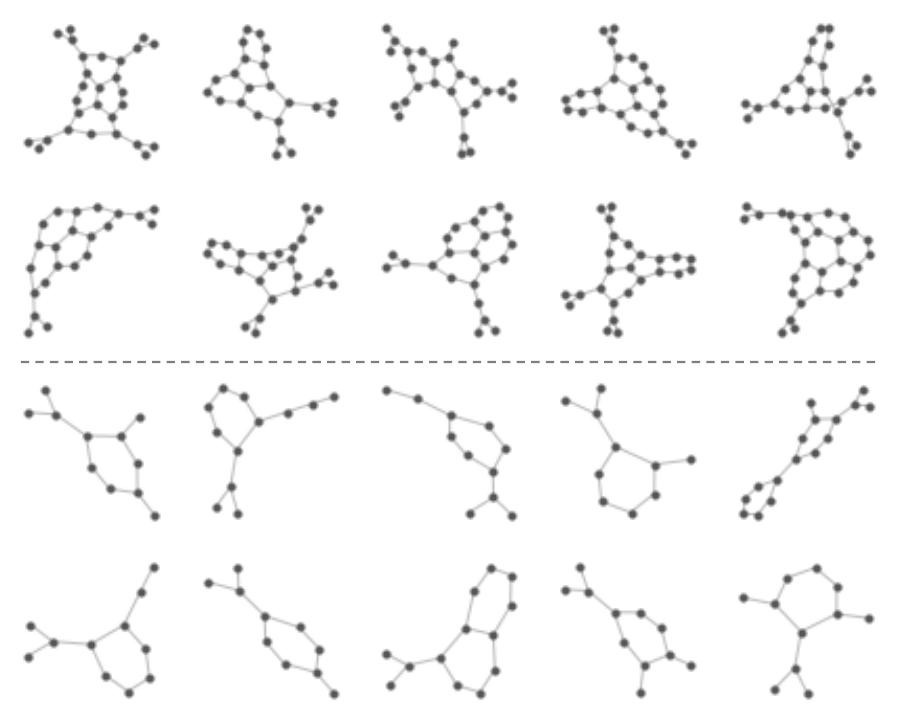}
                \end{center}
                \caption{\CaptionMUTAGgraphs{}}
                \label{fig:MUTAGgraphs}
            \end{wrapfigure}
            of graphs to identify the contrasting explanation, determining what substructure marks the mutagenicity of mutagens, since there are no consistent structural traits defining non-mutagenic molecules. 
            As shown in \autoref{fig:qual}, our method identifies the concurrence of a 6-node ring and the structure of $NO_2$ and $NH_2$ as the contrasting explanation, which aligns the ground truth (\cite{debnath1991structure}).
            In comparison to XGNN and GNNInterpreter, our explanation is the only one that identifies the structure of $NH_2$ and $NO_2$ functional groups. 
            
            Finally, the visualization in \autoref{fig:qual} reveals that nonmutagen graphs also contain carbon rings attached to functional groups, with the primary distinction between classes being the quantity of these substructures. This finding aligns with the mutagenic characteristics reported in \cite{debnath1991structure} and is further supported by the visualizations in our interactive interface, as illustrated in \autoref{fig:MUTAGgraphs}. This underscores the importance of enabling users to explore graph examples and explanations interactively; without such exploration, users might misinterpret the explanations and incorrectly conclude that these substructures are absent in nonmutagen graphs.
           
        \begin{wrapfigure}{r}{0.5\textwidth}
            \begin{center}
            \vspace{-0.4in}
            \includegraphics[width=0.49\textwidth]{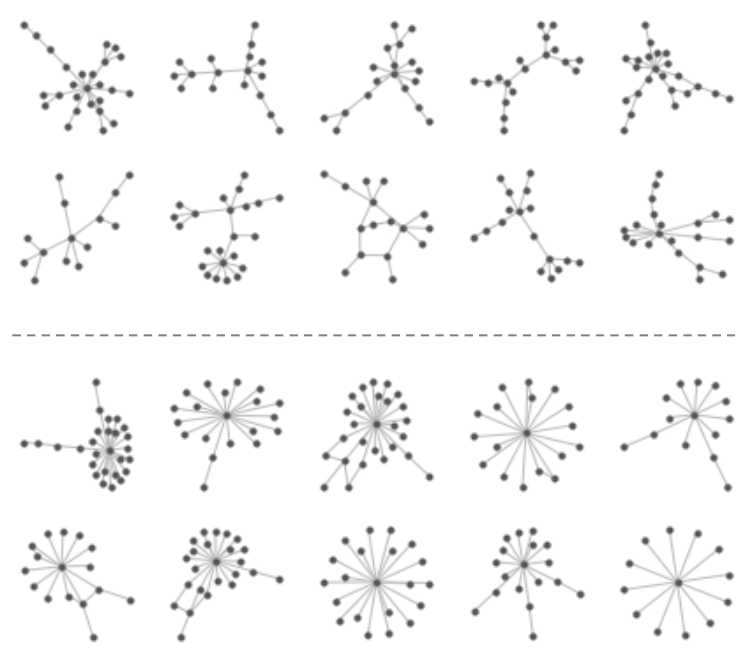}
            \end{center}
            \caption{\CaptionRBmulGrained{}}
            \label{fig:RBmulG}
        \end{wrapfigure}
        \textbf{Reddit-Binary: Non-Trivial Explanations.} 
        \label{sec:RBstudy}
            Explanations for Reddit-Binary graphs are non-trivial and demonstrate the applicability of GNNAnatomy to real-world social network data, as these graphs are neither generated by predefined rules nor labeled based on attached motifs. For the Discussion class, our method identifies a star pattern with an additional branching node, reflecting the branching structures observed in real Discussion graphs. For the Question-Answer (QA) class, it identifies a bipartite structure, where nodes in one set are fully connected to nodes in the other, consistent with QA graphs in which two central nodes connect to all peripheral nodes. 
            
            For the two subsets highlighted in \autoref{fig:projectionMap}, GNNAnatomy identifies the model-level explanation for the Discussion class as the contrasting explanation for the left subset, and the model-level explanation for the QA class for the right subset. This indicates that the branching topology is inconsistently present across Discussion graphs; when absent, the GNN defaults to the QA explanation, which predominates in QA graphs. This interpretation is further supported by the visualizations in our interface, as shown in \autoref{fig:RBmulG}.
    

        \subsection{Evaluations}
        \textbf{Fidelity.}
        \label{sec:fidEval}
            We use the Fidelity (FID) metric (\cite{pope2019explainability, yuan2022explainability, amara2022graphframex}) to measure the change in output probability. To assess explanation sufficiency, we remove edges not involved in any explanatory substructures ($fid-$). To evaluate necessity, we remove edges participating in explanatory graphlets ($fid+$). 
            \begin{equation}
                fid- = \frac{1}{|G_c|}\sum_{g_o \in G_c}||GNN(g_o)[Y_{g_o}]-GNN(g_e)[Y_{g_o}]||_1
            \end{equation}
            \begin{equation}
                fid+ = \frac{1}{|G_c|}\sum_{g_o \in G_c}||GNN(g_o)[Y_{g_o}]-GNN(g_{o \backslash e})[Y_{g_o}]||_1
            \end{equation}
            , where $G_c$ denotes the set of original graphs belonging to class $c$ and $Y_{g_o}$ represents the class label of an original graph $g_o$. The perturbed graph $g_e$ is obtained by removing edges from $g_o$ that are absent from any explanation substructure of class $c$ while $g_{o \backslash e}$ is obtained by removing only the edges participating in any explanation substructure of class $c$. The FID scores are computed as the average $L_1$ distance between the classification probabilities before and after edge removal. The number of removed edges is determined by the Modification Ratio (MR), defined as the proportion of changed edges: $MR = (\#\ of\ removed\ edges) / |E|$, where $E$ is the edge set of the original graph. In summary, a low $fid-$ score indicates the sufficiency of the generated explanations, while a high $fid+$ score reflects their necessity.
            Additionally, since in binary graph-level classification a GNN may recognize a substructure unique to one class while treating the other class as ``something else,'' we apply \autoref{eqt:expIndexGrained} to the entire graph set for the binary classification datasets—Reddit-Binary, MUTAG, and BA-2Motif—to extract global contrasting explanations to evaluate the validity of our multi-grained explanations.
            
\begin{figure}[H]
                \centering
                \includegraphics[width=\linewidth]{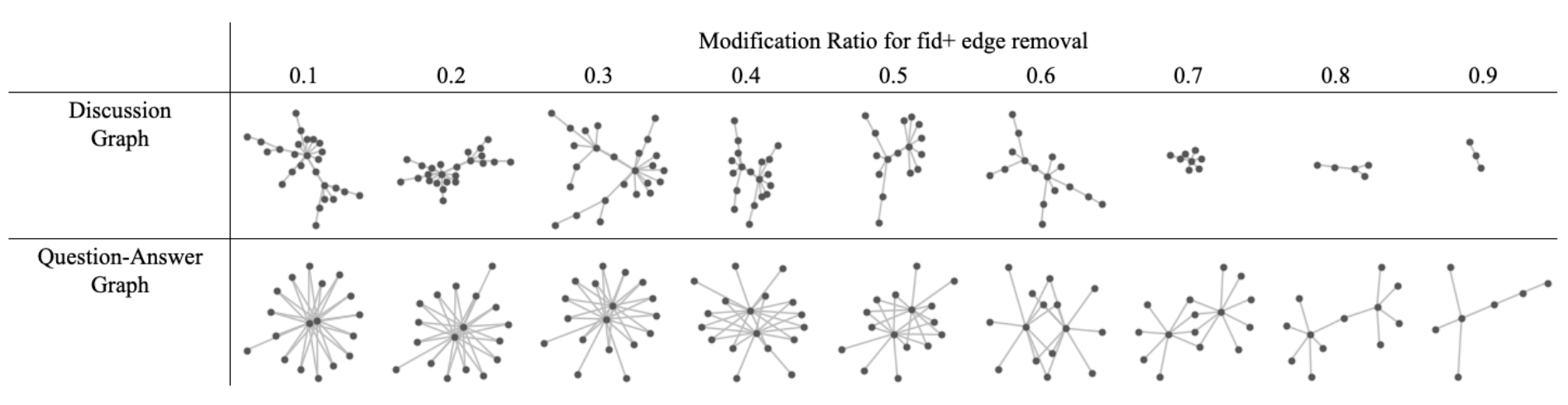}
                \caption{\CaptionRBfid{}}
                \label{fig:RBfid}
            \end{figure}

            As shown in \autoref{tbl:quan}, each of our explanations demonstrates sufficiency or necessity, as indicated by low $fid-$ or high $fid+$ scores, respectively. 
            In Reddit-Binary, the fidelity scores support our identified contrasting explanation for the Discussion class. As shown in \autoref{fig:RBfid}, the pervasive presence of branching substructures in Discussion graphs makes them difficult to remove entirely, even at high MR thresholds ($\geq 0.4$). The low $fid^+$ and $fid^-$ scores suggest that GNNs retain their original classification decisions as long as some branching remains. Furthermore, removing the QA explanatory substructure from QA graphs often introduces branching patterns while preserving parts of the QA structure. The high $fid^+$ score for the QA explanation indicates that the GNN tends to flip its original classification when additional branching substructures are introduced. This strong reliance on the Discussion explanation further validates our identified contrasting explanation.
            
            \vspace{-0.05in}
            \begin{table}[H]
                \caption{\CaptionQuan}
                \label{tbl:quan}
                \includegraphics[width=\linewidth]{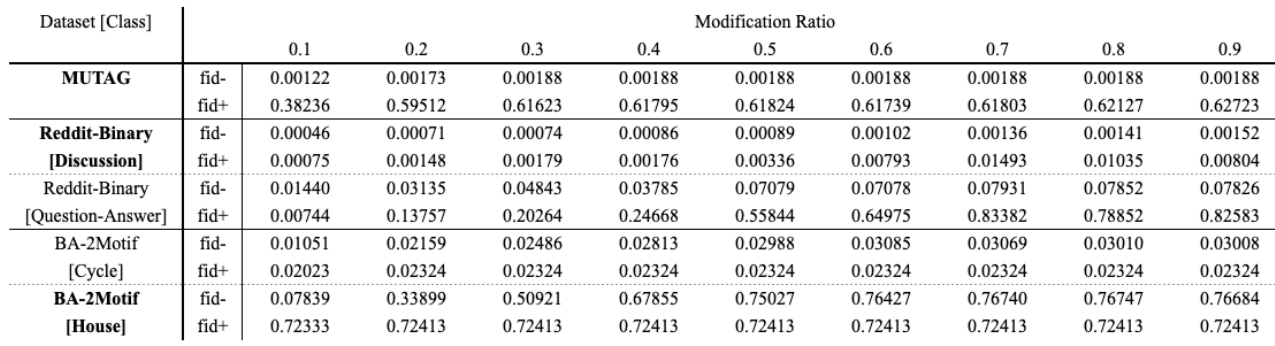}
            \end{table}
            \vspace{-0.25in}

            In BA-2Motif, there is only one explanatory substructure in each graph and removing it does not introduce explanatory substructures from the other class. The high $fid+$ score of the House explanation reinforces our contrasting explanation, suggesting that GNNs primarily rely on the absence of the House explanation for classification, as also highlighted in \autoref{sec:caseStudy}.
    
            \vspace{-0.1in}
            \vspace{-0.15in}
    
            These observations are further supported by visualizations of graphs after perturbation (edge removal), provided in \autoref{fig:RBfid}. As shown in \autoref{fig:RBfid}-top, the ubiquity of this substructure in Discussion graphs makes it difficult to remove entirely, even with a high MR. The low $fid+$ and $fid-$ scores indicate that GNNs can maintain their original classifications as long as this substructure remains. Additionally, removing the Question-Answer (QA) explanatory substructure from QA graphs inadvertently introduces some Discussion explanatory substructures, as shown in \autoref{fig:RBfid}-bottom. The high $fid+$ of the QA explanation reflects GNN’s stronger reliance on the presence of the Discussion explanation for class differentiation.
            Overall, this evaluation further emphasizes the necessity of contrasting explanations for a more comprehensive understanding of GNN behavior.
            
        \textbf{In-Distribution.}
        \label{sec:MMDEval}
            To validate that our explanations remain in-distribution, we computed the Maximum Mean Discrepancy (MMD) using a Gaussian kernel (with $\sigma$ set via the median heuristic) between the distributions of graph embeddings obtained from the trained $GCN_{teacher}$ for both original graphs and their corresponding model-level explanations. These explanations are derived by excluding nodes not involved in any of the identified explanatory substructures.

            As shown in \autoref{tbl:MMD}, the observed MMD scores are consistently near zero, indicating high distributional similarity between the original graph embeddings and their explanations. To assess statistical significance, we conducted a permutation test by computing MMD scores between randomly sampled subsets from the combined set of explanation and original embeddings. We performed ten permutations by randomly shuffling labels and report the resulting permutation-based null MMD range (average and standard deviation).

            In most cases, the observed MMD scores fall below the permutation-based null MMD range (highlighted in red in \autoref{tbl:MMD}), suggesting that the similarity between the original graph and explanation distributions is even stronger than would be expected under the null hypothesis of distributional equivalence.
            However, the Wheel class in the Shape dataset exhibits a larger observed MMD, consistent with our discussion in \autoref{sec:caseStudy}. In this case, the GNN appears to rely on noise artifacts introduced into the Wheel graphs for classification, and the resulting explanation embeddings disproportionately represent these noisy traits, deviating from the original distribution.

            The BA-2Motif House explanations display a modestly elevated MMD yet remain within the null range, confirming statistical indistinguishability. In stark contrast, the Cycle class exhibits an observed MMD that exceeds the permutation-based null range, supporting our hypothesis that the triangle motif dominates the GNN’s learned embeddings. Consequently, any graph lacking triangle substructures is effectively classified as “not a House.” This explains why the Cycle explanation embeddings diverge from the original Cycle graph distribution: the explanations exclude the base graph nodes not involved in the Cycle motif and lack triangle-related structural features in their embeddings. As a result, these embeddings are more arbitrary, reflecting the GNN’s limited ability to meaningfully represent such graphs.

        \vspace{-0.05in}
        \begin{table}[H]
            \caption{\CaptionMMD}
            \label{tbl:MMD}
            \includegraphics[width=\linewidth]{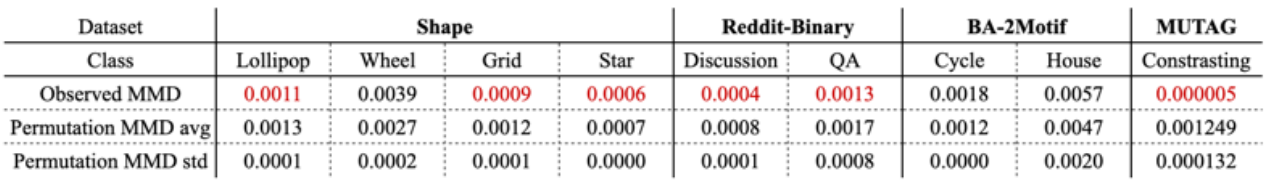}
        \end{table}
        \vspace{-0.25in}
        
        \textbf{Robustness. } 
            To evaluate the robustness of GNNAnatomy, we tested whether it consistently identifies the same explanations under variations in data and model settings. We trained our $MLP_{student}$ on the BA-2Motif dataset ten times with slight experimental changes and identified the global contrasting explanation. Our $MLP_{student}$ consists of a single fully connected layer, with input and output dimensions determined by the size of the graphlet frequency vector and the number of classes, respectively. To introduce noise, we randomly selected 15 out of the 29 graphlets and added the 3-node triangle (identified as the contrasting explanation in \autoref{sec:caseStudy}) as input in each run.

            This setup tests whether the triangle motif remains a key explanatory structure under perturbation. In five runs, GNNAnatomy identified the triangle alone as the contrasting explanation; in the remaining five, it identified a triangle with an additional edge (a "triangle with a tail"). Given that the house motif in BA-2Motif is connected to the Barabási-Albert base graph via an extra edge extending from the triangle, the triangle-with-tail variant carries the same topological interpretation as the triangle alone.
            These results confirm the consistency and robustness of GNNAnatomy in identifying explanations when noise and randomness are introduced.

        \textbf{Alignment. }
        \begin{itemize}[nosep, leftmargin=0.2in]
            \item \textbf{Between Student and Teacher Models.} To assess comprehensively the extent of alignment between the behavior of $GCN_{teacher}$ and $MLP_{student}$, we compute the average mean squared error (MSE), Kullback–Leibler divergence (KL Div), and Cosine similarity (Cos Sim) between their corresponding output classification probabilities for each graph as shown in \autoref{tbl:Alignments}.
            
            \item \textbf{Between PC1 Projection and High-Dimensional Data.} To justify the analogy of proximity in our projection map and the closeness of high-dimensional space, we measure how much variance is retained by the first principal component (PC1) as shown in \autoref{tbl:Alignments}. As suggested in \cite{jolliffe2002principal}, a guideline for variance retention in PCA is between 70-90$\%$, which most datasets achieved, supporting the validity of proximity in the projection map. While Reddit-Binary retained less, \autoref{fig:projectionMap} shows clear separation between classes along the x-axis, indicating that the retained variance encodes vital classification signals. Therefore, the proximity observed in our projection maps correctly reflects the closeness in the higher-dimensional data space.
        \end{itemize}
        
        \vspace{-0.1in}
        \begin{table}[H]
            \caption{\CaptionAlignments}
            \label{tbl:Alignments}
            \includegraphics[width=\linewidth]{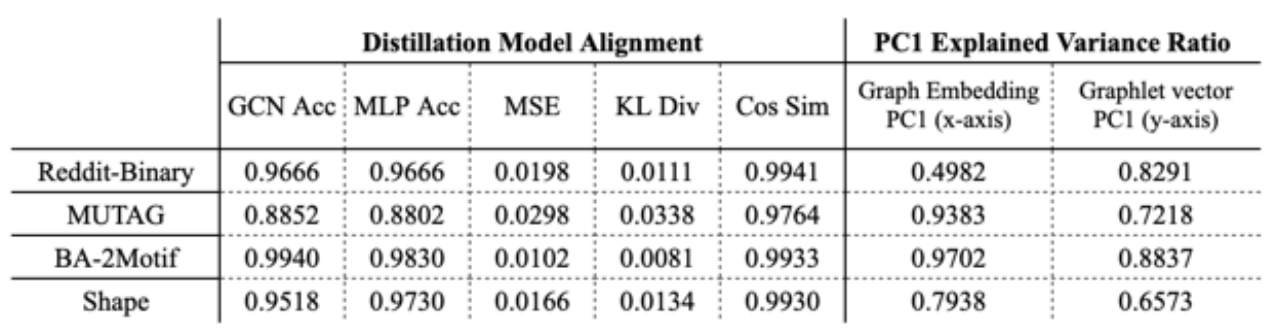}
        \end{table}
        \vspace{-0.25in}

        \textbf{Time Cost Summary.} As discussed in \autoref{sec:background}, graphlet frequency computation is time-consuming but we use Louvain community detection to mitigate the potential scalability issue. Here, we provide a time cost table summarizing the process time for each module in GNNAnatomy.
        As shown in \autoref{tbl:TimeCost}, the graphlet-related computation time is practical and is a one-time processing step per dataset. Only the reported time for Reddit-Binary reflects processing on partitioned graphs; no partitioning was applied to the other datasets. Notably, Reddit-Binary is among the largest graph-level classification datasets in TUDataset (\cite{morris2020tudataset}), yet the processing time remains scalable. The MLP training time is measured when classification accuracy per class surpasses 80$\%$. This time might correlate with the task's difficulty since in MUTAG, it is harder to use only structures to classify molecules, and hence, requires more time.

        \vspace{-0.05in}
        \begin{table}[H]
            \caption{\CaptionTimeCost}
            \label{tbl:TimeCost}
            \includegraphics[width=\linewidth]{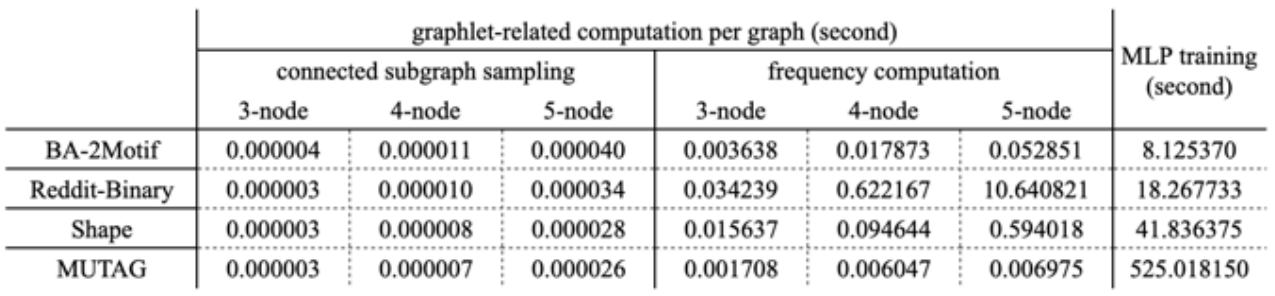}
        \end{table}
        \vspace{-0.25in}

\section{Limitations and Future Works}
    GNNAnatomy currently focuses on providing multi-granular structural explanations for graph classification tasks. Planned future directions include: (1) integrating GraphLime (\cite{huang2022graphlime}) to jointly explain both feature- and structure-based contributions, and (2) extending GNNAnatomy to support node classification.
    For the first direction, our visual analytics interface will be enhanced to display structural explanations from GNNAnatomy alongside feature attributions from GraphLime, enabling users to interpret both types of explanations. For the second, we propose computing node-level graphlet frequency vectors by measuring how often a given node participates in each graphlet type (e.g., how frequently a node appears in instances of $graphlet_{20}$ relative to all such instances in the graph). These node-level graphlet frequency vectors will then serve as input to the transparent $MLP_{student}$ to approximate GNN node classification probabilities. This extension will allow GNNAnatomy to reveal each node’s topological role through interpretable graphlet-based representations.
\section{Conclusion}
We revisit and critically evaluate the foundational assumptions underpinning prior model-level explainable GNN methods, revealing key pitfalls in behavioral alignment, explanation granularity, and evaluability. To overcome these limitations, we introduce GNNAnatomy, a distillation-based framework that explains GNN behavior through transparent, structure-centric reasoning. GNNAnatomy employs graphlets and a lightweight MLP to directly approximate GNN classifications, enabling the identification of class-discriminative and dataset-level contrasting substructures. To support multi-grained insights, we introduce an interactive visual analytics interface that surfaces graph subsets requiring distinct substructures for classification, while also facilitating intuitive, exploratory understanding of GNN decisions. By steering clear of assumption-driven constraints, GNNAnatomy delivers in-distribution, multi-grained, trustworthy, and human-comprehensible explanations that faithfully capture the full behavioral spectrum of GNNs across diverse graph instances.


\section*{Acknowledgments}
\noindent This work was supported by the National Institute of Health through grants 1R01CA270454-01 and 1R01CA273058-01.

\bibliography{00_ref}

\begin{thebibliography}{47}
\providecommand{\natexlab}[1]{#1}
\providecommand{\url}[1]{\texttt{#1}}
\expandafter\ifx\csname urlstyle\endcsname\relax
  \providecommand{\doi}[1]{doi: #1}\else
  \providecommand{\doi}{doi: \begingroup \urlstyle{rm}\Url}\fi

\bibitem[Abbas(2021)]{abbas2021social}
Ash~Mohammad Abbas.
\newblock Social network analysis using deep learning: applications and schemes.
\newblock \emph{Social Network Analysis and Mining}, 11\penalty0 (1):\penalty0 106, 2021.

\bibitem[Amara et~al.(2022)Amara, Ying, Zhang, Han, Shan, Brandes, Schemm, and Zhang]{amara2022graphframex}
Kenza Amara, Rex Ying, Zitao Zhang, Zhihao Han, Yinan Shan, Ulrik Brandes, Sebastian Schemm, and Ce~Zhang.
\newblock Graphframex: Towards systematic evaluation of explainability methods for graph neural networks.
\newblock \emph{arXiv preprint arXiv:2206.09677}, 2022.

\bibitem[Blondel et~al.(2008)Blondel, Guillaume, Lambiotte, and Lefebvre]{blondel2008fast}
Vincent~D Blondel, Jean-Loup Guillaume, Renaud Lambiotte, and Etienne Lefebvre.
\newblock Fast unfolding of communities in large networks.
\newblock \emph{Journal of statistical mechanics: theory and experiment}, 2008\penalty0 (10):\penalty0 P10008, 2008.

\bibitem[Chen et~al.(2024)Chen, Wu, Gupta, and Ying]{chen2024d4explainer}
Jialin Chen, Shirley Wu, Abhijit Gupta, and Rex Ying.
\newblock D4explainer: In-distribution explanations of graph neural network via discrete denoising diffusion.
\newblock \emph{Advances in Neural Information Processing Systems}, 36, 2024.

\bibitem[Debnath et~al.(1991)Debnath, Lopez~de Compadre, Debnath, Shusterman, and Hansch]{debnath1991structure}
Asim~Kumar Debnath, Rosa~L Lopez~de Compadre, Gargi Debnath, Alan~J Shusterman, and Corwin Hansch.
\newblock Structure-activity relationship of mutagenic aromatic and heteroaromatic nitro compounds. correlation with molecular orbital energies and hydrophobicity.
\newblock \emph{Journal of medicinal chemistry}, 34\penalty0 (2):\penalty0 786--797, 1991.

\bibitem[Dinh \& Van~Pham(2021)Dinh and Van~Pham]{dinh2021social}
Xuan~Truong Dinh and Hai Van~Pham.
\newblock Social network analysis based on combining probabilistic models with graph deep learning.
\newblock In \emph{Communication and Intelligent Systems: Proceedings of ICCIS 2020}, pp.\  975--986. Springer, 2021.

\bibitem[Faust(2010)]{faust2010triads}
Katherine Faust.
\newblock A puzzle concerning triads in social networks: Graph constraints and the triad census.
\newblock \emph{Social Networks}, 32\penalty0 (3):\penalty0 221--233, 2010.

\bibitem[Fong \& Vedaldi(2017)Fong and Vedaldi]{fong2017interpretable}
Ruth~C Fong and Andrea Vedaldi.
\newblock Interpretable explanations of black boxes by meaningful perturbation.
\newblock In \emph{Proceedings of the IEEE international conference on computer vision}, pp.\  3429--3437, 2017.

\bibitem[Funke et~al.(2020)Funke, Khosla, and Anand]{funke2020hard}
Thorben Funke, Megha Khosla, and Avishek Anand.
\newblock Hard masking for explaining graph neural networks.
\newblock 2020.

\bibitem[Heimerl et~al.(2020)Heimerl, Kralj, M{\"o}ller, and Gleicher]{heimerl2020embcomp}
Florian Heimerl, Christoph Kralj, Torsten M{\"o}ller, and Michael Gleicher.
\newblock embcomp: Visual interactive comparison of vector embeddings.
\newblock \emph{IEEE Transactions on Visualization and Computer Graphics}, 28\penalty0 (8):\penalty0 2953--2969, 2020.

\bibitem[Hooker et~al.(2019)Hooker, Erhan, Kindermans, and Kim]{hooker2019benchmark}
Sara Hooker, Dumitru Erhan, Pieter-Jan Kindermans, and Been Kim.
\newblock A benchmark for interpretability methods in deep neural networks.
\newblock \emph{Advances in neural information processing systems}, 32, 2019.

\bibitem[Huang et~al.(2022)Huang, Yamada, Tian, Singh, and Chang]{huang2022graphlime}
Qiang Huang, Makoto Yamada, Yuan Tian, Dinesh Singh, and Yi~Chang.
\newblock Graphlime: Local interpretable model explanations for graph neural networks.
\newblock \emph{IEEE Transactions on Knowledge and Data Engineering}, 2022.

\bibitem[Jin et~al.(2021)Jin, Zeng, Xia, Huang, and Liu]{jin2021application}
Shuting Jin, Xiangxiang Zeng, Feng Xia, Wei Huang, and Xiangrong Liu.
\newblock Application of deep learning methods in biological networks.
\newblock \emph{Briefings in bioinformatics}, 22\penalty0 (2):\penalty0 1902--1917, 2021.

\bibitem[Jin et~al.(2022)Jin, Wang, Wang, Ming, Ma, and Qu]{jin2022gnnlens}
Zhihua Jin, Yong Wang, Qianwen Wang, Yao Ming, Tengfei Ma, and Huamin Qu.
\newblock Gnnlens: A visual analytics approach for prediction error diagnosis of graph neural networks.
\newblock \emph{IEEE Transactions on Visualization and Computer Graphics}, 2022.

\bibitem[Jolliffe(2002)]{jolliffe2002principal}
Ian~T Jolliffe.
\newblock \emph{Principal component analysis for special types of data}.
\newblock Springer, 2002.

\bibitem[Kipf \& Welling(2016)Kipf and Welling]{kipf2016semi}
Thomas~N Kipf and Max Welling.
\newblock Semi-supervised classification with graph convolutional networks.
\newblock \emph{arXiv preprint arXiv:1609.02907}, 2016.

\bibitem[Kwon et~al.(2018)Kwon, Crnovrsanin, and Ma]{Kwon2018}
Oh{-}Hyun Kwon, Tarik Crnovrsanin, and Kwan{-}Liu Ma.
\newblock What would a graph look like in this layout? {A} machine learning approach to large graph visualization.
\newblock \emph{{IEEE} Trans. Vis. Comput. Graph.}, 24\penalty0 (1):\penalty0 478--488, 2018.
\newblock \doi{10.1109/TVCG.2017.2743858}.
\newblock URL \url{https://doi.org/10.1109/TVCG.2017.2743858}.

\bibitem[La~Rosa et~al.(2023)La~Rosa, Blasilli, Bourqui, Auber, Santucci, Capobianco, Bertini, Giot, and Angelini]{la2023state}
Biagio La~Rosa, Graziano Blasilli, R~Bourqui, D~Auber, Giuseppe Santucci, Roberto Capobianco, Enrico Bertini, Romain Giot, and Marco Angelini.
\newblock State of the art of visual analytics for explainable deep learning.
\newblock In \emph{Computer Graphics Forum}, volume~42, pp.\  319--355. Wiley Online Library, 2023.

\bibitem[Li et~al.(2018)Li, Njotoprawiro, Haleem, Chen, Yi, and Ma]{li2018embeddingvis}
Quan Li, Kristanto~Sean Njotoprawiro, Hammad Haleem, Qiaoan Chen, Chris Yi, and Xiaojuan Ma.
\newblock Embeddingvis: A visual analytics approach to comparative network embedding inspection.
\newblock In \emph{2018 IEEE Conference on Visual Analytics Science and Technology (VAST)}, pp.\  48--59. IEEE, 2018.

\bibitem[Li et~al.(2021)Li, Yuan, Radfar, Marendy, Ni, O’Brien, and Casillas-Espinosa]{li2021graph}
Rui Li, Xin Yuan, Mohsen Radfar, Peter Marendy, Wei Ni, Terrence~J O’Brien, and Pablo~M Casillas-Espinosa.
\newblock Graph signal processing, graph neural network and graph learning on biological data: a systematic review.
\newblock \emph{IEEE Reviews in Biomedical Engineering}, 16:\penalty0 109--135, 2021.

\bibitem[Liu et~al.(2022)Liu, Wang, Bernard, and Munzner]{liu2022visualizing}
Zipeng Liu, Yang Wang, J{\"u}rgen Bernard, and Tamara Munzner.
\newblock Visualizing graph neural networks with corgie: Corresponding a graph to its embedding.
\newblock \emph{IEEE Transactions on Visualization and Computer Graphics}, 28\penalty0 (6):\penalty0 2500--2516, 2022.

\bibitem[Lundberg(2017)]{lundberg2017unified}
Scott Lundberg.
\newblock A unified approach to interpreting model predictions.
\newblock \emph{arXiv preprint arXiv:1705.07874}, 2017.

\bibitem[Luo et~al.(2020)Luo, Cheng, Xu, Yu, Zong, Chen, and Zhang]{luo2020parameterized}
Dongsheng Luo, Wei Cheng, Dongkuan Xu, Wenchao Yu, Bo~Zong, Haifeng Chen, and Xiang Zhang.
\newblock Parameterized explainer for graph neural network.
\newblock \emph{Advances in neural information processing systems}, 33:\penalty0 19620--19631, 2020.

\bibitem[Min et~al.(2021)Min, Gao, Peng, Wang, Qin, and Fang]{min2021stgsn}
Shengjie Min, Zhan Gao, Jing Peng, Liang Wang, Ke~Qin, and Bo~Fang.
\newblock Stgsn—a spatial--temporal graph neural network framework for time-evolving social networks.
\newblock \emph{Knowledge-Based Systems}, 214:\penalty0 106746, 2021.

\bibitem[Morris et~al.(2020)Morris, Kriege, Bause, Kersting, Mutzel, and Neumann]{morris2020tudataset}
Christopher Morris, Nils~M Kriege, Franka Bause, Kristian Kersting, Petra Mutzel, and Marion Neumann.
\newblock Tudataset: A collection of benchmark datasets for learning with graphs.
\newblock \emph{arXiv preprint arXiv:2007.08663}, 2020.

\bibitem[Muzio et~al.(2021)Muzio, O’Bray, and Borgwardt]{muzio2021biological}
Giulia Muzio, Leslie O’Bray, and Karsten Borgwardt.
\newblock Biological network analysis with deep learning.
\newblock \emph{Briefings in bioinformatics}, 22\penalty0 (2):\penalty0 1515--1530, 2021.

\bibitem[Pereira et~al.(2023)Pereira, Nascimento, Resck, Mesquita, and Souza]{pereira2023distill}
Tamara Pereira, Erik Nascimento, Lucas~E Resck, Diego Mesquita, and Amauri Souza.
\newblock Distill n’explain: explaining graph neural networks using simple surrogates.
\newblock In \emph{International Conference on Artificial Intelligence and Statistics}, pp.\  6199--6214. PMLR, 2023.

\bibitem[Pope et~al.(2019)Pope, Kolouri, Rostami, Martin, and Hoffmann]{pope2019explainability}
Phillip~E Pope, Soheil Kolouri, Mohammad Rostami, Charles~E Martin, and Heiko Hoffmann.
\newblock Explainability methods for graph convolutional neural networks.
\newblock In \emph{Proceedings of the IEEE/CVF conference on computer vision and pattern recognition}, pp.\  10772--10781, 2019.

\bibitem[Pr{\v{z}}ulj(2007)]{prvzulj2007biological}
Nata{\v{s}}a Pr{\v{z}}ulj.
\newblock Biological network comparison using graphlet degree distribution.
\newblock \emph{Bioinformatics}, 23\penalty0 (2):\penalty0 e177--e183, 2007.

\bibitem[Ribeiro et~al.(2016)Ribeiro, Singh, and Guestrin]{ribeiro2016should}
Marco~Tulio Ribeiro, Sameer Singh, and Carlos Guestrin.
\newblock " why should i trust you?" explaining the predictions of any classifier.
\newblock In \emph{Proceedings of the 22nd ACM SIGKDD international conference on knowledge discovery and data mining}, pp.\  1135--1144, 2016.

\bibitem[Sanchez-Lengeling et~al.(2020)Sanchez-Lengeling, Wei, Lee, Reif, Wang, Qian, McCloskey, Colwell, and Wiltschko]{sanchez2020evaluating}
Benjamin Sanchez-Lengeling, Jennifer Wei, Brian Lee, Emily Reif, Peter Wang, Wesley Qian, Kevin McCloskey, Lucy Colwell, and Alexander Wiltschko.
\newblock Evaluating attribution for graph neural networks.
\newblock \emph{Advances in neural information processing systems}, 33:\penalty0 5898--5910, 2020.

\bibitem[Schlichtkrull et~al.(2021)Schlichtkrull, Cao, and Titov]{schlichtkrull2021interpreting}
Michael~Sejr Schlichtkrull, Nicola~De Cao, and Ivan Titov.
\newblock Interpreting graph neural networks for {\{}nlp{\}} with differentiable edge masking.
\newblock In \emph{International Conference on Learning Representations}, 2021.
\newblock URL \url{https://openreview.net/forum?id=WznmQa42ZAx}.

\bibitem[Tan et~al.(2019)Tan, Liu, and Hu]{tan2019deep}
Qiaoyu Tan, Ninghao Liu, and Xia Hu.
\newblock Deep representation learning for social network analysis.
\newblock \emph{Frontiers in big Data}, 2:\penalty0 2, 2019.

\bibitem[Ugander et~al.(2013)Ugander, Backstrom, and Kleinberg]{Ugander}
J.~Ugander, L.~Backstrom, and J.~Kleinberg.
\newblock Subgraph frequencies: Mapping the empirical and extremal geography of large graph collections.
\newblock In \emph{Proceedings of the International Conference on World Wide Web}, pp.\  1307–1318, 2013.

\bibitem[Vu \& Thai(2020)Vu and Thai]{vu2020pgm}
Minh Vu and My~T Thai.
\newblock Pgm-explainer: Probabilistic graphical model explanations for graph neural networks.
\newblock \emph{Advances in neural information processing systems}, 33:\penalty0 12225--12235, 2020.

\bibitem[Wang et~al.(2024)Wang, Liu, and Zhang]{wang2024visual}
Junpeng Wang, Shixia Liu, and Wei Zhang.
\newblock Visual analytics for machine learning: A data perspective survey.
\newblock \emph{IEEE Transactions on Visualization and Computer Graphics}, 2024.

\bibitem[Wang et~al.(2020)Wang, Wu, Zhang, He, and Chua]{wang2020causal}
Xiang Wang, Yingxin Wu, An~Zhang, Xiangnan He, and Tat-seng Chua.
\newblock Causal screening to interpret graph neural networks.
\newblock 2020.

\bibitem[Wang et~al.(2021)Wang, Wu, Zhang, He, and Chua]{wang2021towards}
Xiang Wang, Yingxin Wu, An~Zhang, Xiangnan He, and Tat-Seng Chua.
\newblock Towards multi-grained explainability for graph neural networks.
\newblock \emph{Advances in Neural Information Processing Systems}, 34:\penalty0 18446--18458, 2021.

\bibitem[Wang \& Shen(2022)Wang and Shen]{wang2022gnninterpreter}
Xiaoqi Wang and Han-Wei Shen.
\newblock Gnninterpreter: A probabilistic genereative model-level explanation for graph neural networks.
\newblock \emph{arXiv preprint arXiv:2209.07924}, 2022.

\bibitem[Yanardag \& Vishwanathan(2015)Yanardag and Vishwanathan]{yanardag2015deep}
Pinar Yanardag and SVN Vishwanathan.
\newblock Deep graph kernels.
\newblock In \emph{Proceedings of the 21th ACM SIGKDD international conference on knowledge discovery and data mining}, pp.\  1365--1374, 2015.

\bibitem[Yin et~al.(2023)Yin, Li, Yan, Lian, and Wang]{yin2023train}
Jun Yin, Chaozhuo Li, Hao Yan, Jianxun Lian, and Senzhang Wang.
\newblock Train once and explain everywhere: Pre-training interpretable graph neural networks.
\newblock \emph{Advances in Neural Information Processing Systems}, 36:\penalty0 35277--35299, 2023.

\bibitem[Ying et~al.(2018)Ying, You, Morris, Ren, Hamilton, and Leskovec]{ying2018hierarchical}
Zhitao Ying, Jiaxuan You, Christopher Morris, Xiang Ren, Will Hamilton, and Jure Leskovec.
\newblock Hierarchical graph representation learning with differentiable pooling.
\newblock \emph{Advances in neural information processing systems}, 31, 2018.

\bibitem[Ying et~al.(2019)Ying, Bourgeois, You, Zitnik, and Leskovec]{ying2019gnnexplainer}
Zhitao Ying, Dylan Bourgeois, Jiaxuan You, Marinka Zitnik, and Jure Leskovec.
\newblock {GNNExplainer}: Generating explanations for graph neural networks.
\newblock \emph{Advances in neural information processing systems}, 32, 2019.

\bibitem[Yuan et~al.(2020)Yuan, Tang, Hu, and Ji]{yuan2020xgnn}
Hao Yuan, Jiliang Tang, Xia Hu, and Shuiwang Ji.
\newblock Xgnn: Towards model-level explanations of graph neural networks.
\newblock In \emph{Proceedings of the 26th ACM SIGKDD International Conference on Knowledge Discovery \& Data Mining}, pp.\  430--438, 2020.

\bibitem[Yuan et~al.(2021)Yuan, Yu, Wang, Li, and Ji]{yuan2021explainability}
Hao Yuan, Haiyang Yu, Jie Wang, Kang Li, and Shuiwang Ji.
\newblock On explainability of graph neural networks via subgraph explorations.
\newblock In \emph{International conference on machine learning}, pp.\  12241--12252. PMLR, 2021.

\bibitem[Yuan et~al.(2022)Yuan, Yu, Gui, and Ji]{yuan2022explainability}
Hao Yuan, Haiyang Yu, Shurui Gui, and Shuiwang Ji.
\newblock Explainability in graph neural networks: A taxonomic survey.
\newblock \emph{IEEE Transactions on Pattern Analysis and Machine Intelligence}, 2022.

\bibitem[Zhang et~al.(2021)Zhang, Defazio, and Ramesh]{zhang2021relex}
Yue Zhang, David Defazio, and Arti Ramesh.
\newblock Relex: A model-agnostic relational model explainer.
\newblock In \emph{Proceedings of the 2021 AAAI/ACM Conference on AI, Ethics, and Society}, pp.\  1042--1049, 2021.

\end{thebibliography}
\bibliographystyle{tmlr}

\appendix
\newpage
\section*{Appendix}
\vspace{0.1in}
\section{Visual Analytics Interface}
    \begin{figure*}[!ht]
        \centering
        \includegraphics[width=\linewidth]{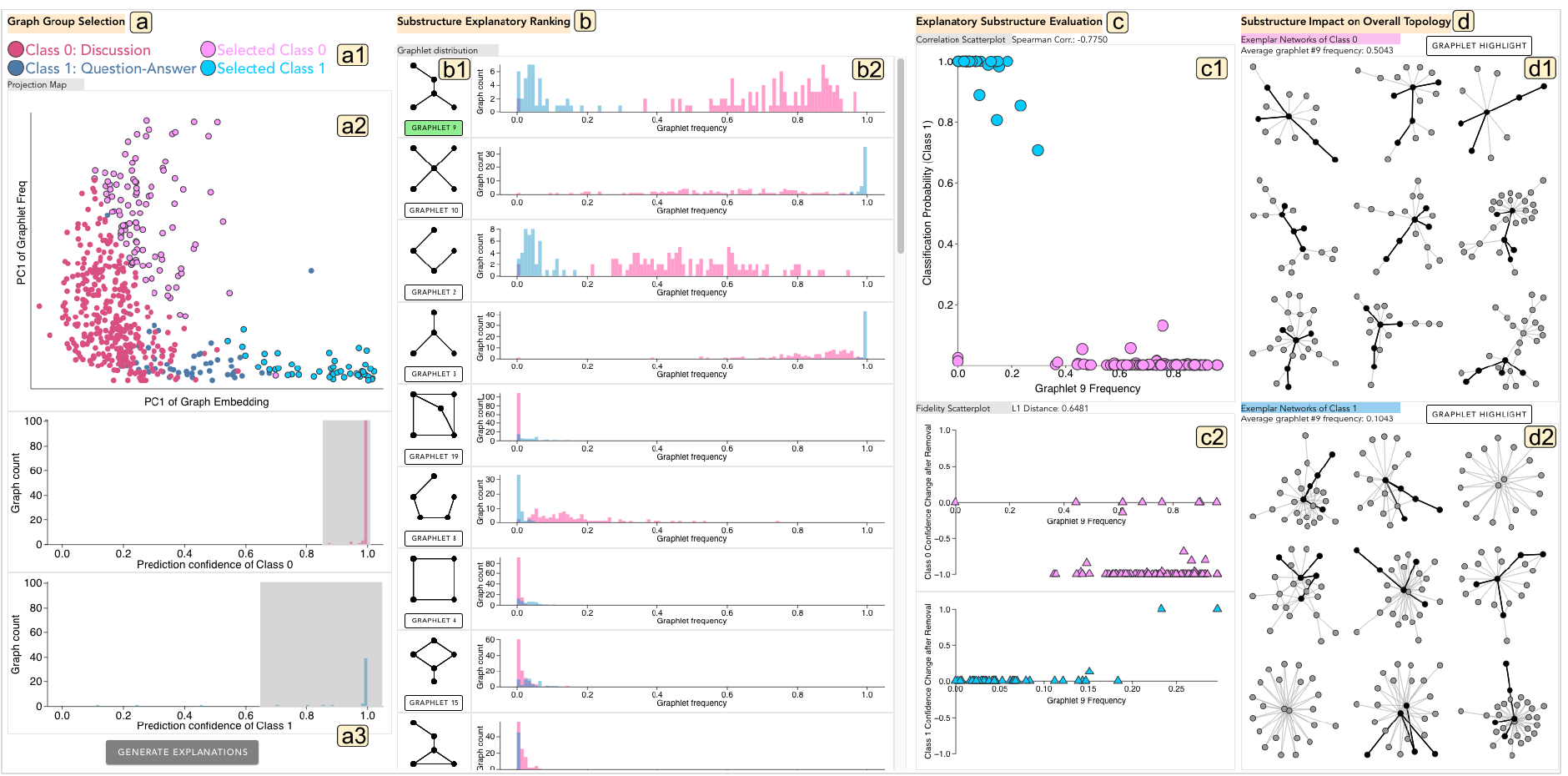}
        \caption{\CaptionUI}
        \label{fig:ui}
    \end{figure*}

    Our visual analytics system is a pivotal tool for interactively developing and evaluating explanations. It consists of four columns that facilitate the GNNAnatomy workflow, as shown in \autoref{fig:ui}. Below, we provide a brief summary of the interface’s key functions and how they contribute to the overall workflow.
    \begin{enumerate}[]
        \item Beginning with the \textit{Graph Group Selection} column (\autoref{fig:ui}(a)), the \textit{projection map} (\autoref{fig:ui}(a2)) helps users identify graph groups that exhibit incoherent topological traits, which require different substructures for differentiating classes within each group. The selection made in \autoref{fig:ui}(a2) can be further refined in \autoref{fig:ui}(a3) based on the class and classification confidence of the selected graphs. 
        \item After selecting a group of graphs in \autoref{fig:ui}(a), the \textit{Substructure Explanatory Ranking} column (\autoref{fig:ui}(b)) depicts each graphlet (\autoref{fig:ui}(b1)), ranked based on value of contrasting activation introduced in \autoref{eqt:expIndexGrained}. It also shows the class-wise distribution of graphs based on the frequency of each graphlet (\autoref{fig:ui}(b2)).
        \item Upon selecting a graphlet in \autoref{fig:ui}(b), the \textit{Explanatory Substructure Evaluation} column (\autoref{fig:ui}(c)) reveals the correlation between the activations of the chosen graphlet and the GNN's classification probability (\autoref{fig:ui}(c1)). \autoref{fig:ui}(c2) provides class-wise scatterplots illustrating changes in classification confidence when the selected graphlet is removed. Based on these patterns, users can then choose specific graphs to further examine their topology in \autoref{fig:ui}(d).
        \item The \textit{Substructure Impact on Overall Topology} column (\autoref{fig:ui}(d)) illustrates how the selected graphlet is exhibited in the overall graph topology. A set of representative graphs from the two classes chosen in \autoref{fig:ui}(c) is displayed (\autoref{fig:ui}(d1, d1)). When the ``graphlet highlight'' button is clicked, the selected graphlet is highlighted in each graph where it appears. Clicking again highlights the next graphlet.
    \end{enumerate}

    The above walk-through illustrates a scenario of explaining GNN's behavior on datasets with binary categorical graph labels. For datasets with non-binary labels, the visualization helps identify the appropriate level of explanation for two selected classes at a time. If no user-specified level is provided in the Projection Map, GNNAnatomy defaults to generating model-level explanations for each class, regardless of the number of classes in the dataset.

    In this interface, the probability output for the class to which a graph belongs is referred to as \textit{classification confidence}, while the probability output specifically for $Class_1$ is termed \textit{classification probability}. For instance, if a Discussion graph (i.e., $Class_0$) has a GNN probability output of $[0.75, 0.25]$, its classification confidence is $0.75$ and its classification probability is $0.25$. In the following paragraphs, we use this terminology to introduce each visual component in detail, highlighting the information each conveys.

    \subsection{Graph Group Selection}
        As shown in \autoref{fig:ui}(a), this column consists of three components. First, \autoref{fig:ui}(a1) displays the color legends used throughout the interface. Next, \autoref{fig:ui}(a2) features the \textit{Projection Map}, which projects all graphs from the analyzed dataset. The x-axis represents the first principal component (PC1) of the 80D graph embedding generated by the trained GNN. The proximity along the x-axis indicates how similarly the GNN perceives the graphs; graphs closer together are considered more similar in terms of the topological traits captured to generate their embeddings. The y-axis represents the PC1 of the 29D graphlet frequency vectors, with proximity along this axis reflecting the similarity in graph topologies as characterized by the graphlets. Graphs closer on the y-axis have more similar topologies.
        
        The relationship between these two sets of projections helps users identify graph groups that require distinct topological traits for class differentiation, as detailed in \autoref{sec:method_PM}.
        Guided by the associations revealed in the projection map, users can lasso-select a group of graphs for which explanations are generated, highlighted with pink and cyan outlines. \autoref{fig:ui}(a3) allows for further refinement of this selection. The chosen graphs are displayed in two histograms, sorted by class label, showing their distribution across GNN-generated confidence scores. This refinement feature enables users to brush and select specific sections of confidence scores by class.
        In summary, this column, specifically the projection map, enables the following generation of GNN explanations at flexible levels for required granularity.
        
    \subsection{Substructure Explanatory Ranking}
        As shown in \autoref{fig:ui}(b), each graphlet is depicted in \autoref{fig:ui}(b1) with a histogram in \autoref{fig:ui}(b2) presenting the class-wise distribution of graphs by frequency. The graphlet images provide an intuitive understanding of the substructures, while the histogram reveals differences in frequency domains across classes, helping users compare graphlet substructures and select one for further validation of its relevance to GNN behavior.        
        The ranking of graphlets (i.e., the order from top to bottom) is based on the contrasting activation value as introduced in \autoref{eqt:expIndexGrained}. 
        
    \subsection{Explanatory Substructure Evaluation}
        As shown in \autoref{fig:ui}(c), we provide two visualizations: the correlation scatterplot in \autoref{fig:ui}(c1) and the fidelity scatterplot in \autoref{fig:ui}(c2).
        
        \textbf{Correlation scatterplot. }
        In \autoref{fig:ui}(c1), each graph in the selected group is plotted with the activation of the chosen graphlet on the x-axis and the classification probability on the y-axis. This scatterplot helps users examine how well the activation of the chosen graphlet aligns with GNN behavior for class differentiation. The Spearman's correlation coefficient, displayed above the scatterplot, provides a quantitative measure of this relationship.

        \textbf{Fidelity scatterplot. }
        In \autoref{fig:ui}(c2), each triangle represents a graph. The two scatterplots show the changes in GNN's classification confidence after removing the chosen graphlet from graphs of each class. The scatterplots differentiate between positive effects (improved confidence) and negative effects (decreased confidence) along the y-axis. The x-axis represents the frequencies of the selected graphlet in the original graphs. Additionally, the average change, calculated as the mean absolute L1 distance between original and perturbed classification confidences, is displayed above scatterplots, summarizing the impact of graphlet removal.

    \subsection{Substructure Impact on Overall Topology} 
        Guided by the correlation patterns in \autoref{fig:ui}(c1), users select graphs to analyze how the chosen graphlet substructure is exhibited in the overall graph topology. In \autoref{fig:ui}(d), two sets of graph visualizations are provided: one for representative graphs of $Class_0$ and the other for $Class_1$. 
        For each class, if more than nine graphs are selected, nine are randomly picked among the selected. Displaying an ensemble of nine graphs per class allows users to explore a variety of graph structures within the selected range while preserving visual clarity. 
        The average frequency of the chosen graphlet is displayed above the visualizations to aid in interpretation. Clicking the ``graphlet highlight'' button will highlight one instance of the selected graphlet substructure in opaque black. Since each graph's topology may contain multiple substructures, clicking the button again will highlight the next matching substructure, if available. This approach helps users understand how the substructural impact translates to overall topology differences and reason about the semantics of the explanatory graphlet, aided by the graph labels.

\section{Comparison to Other Multi-Grained Approaches}
    Similar to the concepts in ReFine and $\pi$-GNN (\cite{wang2021towards, yin2023train}), $MLP_{student}$ learns the knowledge space of GNNs that can be globally applied to all graphs. However, ReFine lacks a clear mapping to actual structural representations, limiting the interpretability of its explanations. Additionally, $\pi$-GNN requires ground truth for training and both methods finetune the learned global knowledge to each instance. In contrast, GNNAnatomy eliminates the need for further training or ground truth while enabling flexible generation at the required level (i.e., any graph subset identified from our Projection Map) of an interpretable finer-grained explanation (i.e., the most contrasting substructure between classes).
    
\section{Complete List of Graphlets}
    \begin{figure}[H]
        \centering
        \includegraphics[width=\linewidth]{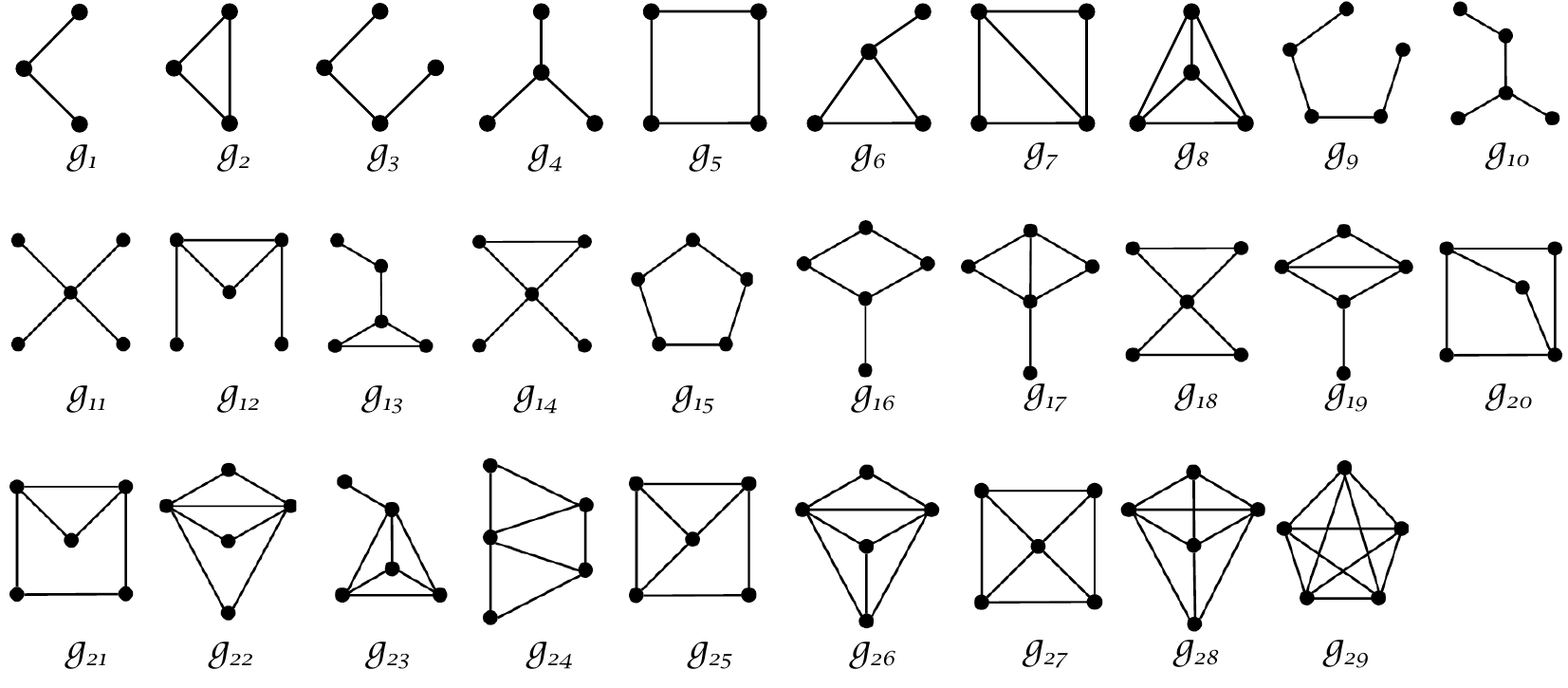}
        \caption{\CaptionGraphletFull{}}
    \end{figure}

\end{document}